\documentclass[conference,compsoc]{IEEEtran}
\usepackage{times}
\usepackage{epsfig}
\usepackage{graphicx}
\usepackage{amsmath}
\usepackage{amssymb}
\usepackage{dsfont}
\usepackage{epsfig}
\usepackage{xspace}
\usepackage{graphicx}
\usepackage{epstopdf}
\usepackage{subfig}
\usepackage{multirow}

\ifCLASSOPTIONcompsoc
  \usepackage[nocompress]{cite}
\else
  \usepackage{cite}
\fi
\ifCLASSINFOpdf
\else
\fi
\begin{document}
\title{Zero Shot Hashing}

\author{\IEEEauthorblockN{Shubham Pachori}
\IEEEauthorblockA{Electrical Engineering\\
Indian Institute of Technology Gandhinagar\\
Gandhinagar, Gujarat 382355\\
Email: shubham\_pachori@iitgn.ac.in}
\and
\IEEEauthorblockN{Shanmuganathan Raman}
\IEEEauthorblockA{Electrical Engineering \& \\Computer Science and Engineering\\
Indian Institute of Technology Gandhinagar\\
Gandhinagar, Gujarat 382355\\
Email: shanmuga@iitgn.ac.in}}


% compsoc conferences compared to traditional conferences), use this
%Telephone: (800) 555--1212, Fax: (888) 555--1212}


\maketitle

\begin{abstract}
This paper provides a framework to hash images containing instances of unknown object classes. In many object recognition problems, we might have access to huge amount of data. It may so happen that even this huge data doesn't cover the objects belonging to classes that we see in our day to day life. Zero shot learning exploits auxiliary information (also called as signatures) in order to predict the labels corresponding to unknown classes. In this work, we attempt to generate the hash codes for images belonging to unseen classes, information of which is available only through the textual corpus. We  formulate this as an unsupervised hashing formulation as the exact labels are not available for the instances of unseen classes. We show that the proposed solution is able to generate hash codes which can predict labels corresponding to unseen classes with appreciably good precision. 
\end{abstract}


% \fi

\section{Introduction}

With billions of image-based data information added to social networking sites like Flickr, Instagram everday, it has become challenge to accurately organize the data. One possible solution is to use hashing techniques with the smallest number of possible bits in order to reduce both storage requirement and query response time. The hashes correponding to images thus obtained can be used for multiple computer vision tasks such as recogniton, image retrieval and understanding. Hashing based approximate nearest neighbor(ANN) search methods have attracted a lot of attention in the past two decades. Apart from this, considerable amount of research has been done in the past few years on improving zero shot learning algorithms [\cite{palatucci2009zero}, \cite{socher2013zero}, \cite{kodirov2015unsupervised}, \cite{elhoseiny2013write}, \cite{lei2015predicting}, \cite{zhang2015zero}].
Zero shot learning requires one to transfer knowledge learnt from the classes present in the training data to the classes which are not been observed yet. This knowledge is generally available in the form of signatures or attributes along with visual concept. 
Our main motivation for zero shot hashing (referred as ZSH in the rest of the paper) comes from the fact that humans learn to visualize an image from just the attributes present in it. For example, if we are given an  dictionary where we store images, then we are likely to keep the image of liger in between the images of tiger and lion \cite{dawkins2012illustrated}. This has also been shown by Yosinski \cite{yosinski2015understanding}, where each filter present in CNN during training learns to detect different features in an image like clothes, face, numbers, etc. This cumulative knowledge helps to determine the classes to which the given image belongs to. 
The concept of attribute based learning has also been used in multiple instance learning, where bag of instances with same labels are created. These instances contain different attributes of an image such as strips, ears of leopard etc.\\  
Features learnt from other modalities like text are transferred inductively to images using experience. The features extraction methods like CNN contain infromation about all the attributes present in a given image. However, they do not contain information about other modalities like text. Thus, we could keep an image of liger in between tiger and lion but we are allowed to create another class called "LIGER" in our dictionary. Moreover, it has been shown that using supervised information of an image along with its features could significantly improve the hash codes of image \cite{xia2014supervised}, \cite{liu2012supervised}, \cite{norouzi2011minimal}.\\

The primary contributions of this paper are listed below:
\begin{enumerate}
  \item Incorporating zero shot learning framework in induction based unsupervised hashing problem.
  \item Learning correspondence between signatures of classes and images while jointly embedding them into a common space.
  \item A novel but a simple approach to address out-of-sample extension problems associated with hashing images belonging to instances of unseen categories. 
\end{enumerate}

The outline of this paper is as follows. In section 2, we discuss the related works done previously and why our method is different from them. In section 3, we discuss the proposed methodology. In section 4, we evaluate our method and discuss the results obtained from our experiments. In section 5, we conclude our paper along with works in future. 

\section{Related Work}
Many hashing methods have been proposed in the past, which can be categorized into two type - data dependent and data independent hashing. Locality sensitive hashing is the most popular data independent hashing technique, which uses randomized projections to generate hash functions and ensures high collision probability for similar data points. Variants of LSH \cite{gionis1999similarity} have been developed by taking different distance measures like Mahalanobis distance \cite{kulis2009fast}, kernel similarity (\cite{kulis2009kernelized},\cite{raginsky2009locality}) and $p-$norm distance \cite{datar2004locality}. Other forms of LSH could be found in detail in \cite{wang2014hashing}. In general data independent hashing techniques exploit long hashes and several hash tables to achieve better performance in terms of precision and recall, thus rendering them limited in use for large scale applications. On the contrary, data dependent techniques could be classified into two types - supervised and unsupervised hashing. These algorithms tend to exploit the available training data to generate short binary codes. Among the data dependent methods, PCA-based hashing (\cite{gong2013iterative}, \cite{wang2012semi}), supervised and semi-supervised hashing (\cite{wang2012semi}, \cite{liu2012supervised}, \cite{norouzi2011minimal}, \cite{kulis2009learning}) and graph based hashing (\cite{liu2011hashing}, \cite{weiss2009spectral}) techniques are quite popular.  

It has been shown that leveraging non-linear manifold embedding techniques have helped in generating better pairwise affinity preserving dense binary codes. Among well-known hashing algorithms which utilize this idea is spectral hashing (\cite{weiss2009spectral}, \cite{weiss2012multidimensional}), which uses the eigenfunctions of the Laplacian matrix to capture variation in the data. anchor graph hashing, popularly known as AGH (\cite{liu2011hashing}, \cite{liu2014discrete}), uses anchor graphs for generating hash codes for training and out-of-sample data efficiently and effectively. Benefiting from the properties of manifold approaches, Inductive manifold hashing (IMH) has been proposed in \cite{shen2013inductive}, which computes the manifold of a given data point according to the manifold of its neighbours. In \cite{shen2015learning}, a solution has been proposed for preserving the inner-product similarities among raw vectors, while tackling maximum inner product search (MIPS) problem.

Apart from these, extensive research has been done in producing multimodal hash codes. Data available to us is in multiple information types and contains both text tags and visual concepts. The concept behind cross modal hashing (CMH) methods, in general, is to project the multimodal data in a common hamming space so that the distance between similar data in heterogeneous modalities are preserved. In \cite{zhu2013linear}, linear cross-modal hashing (LCMH) has been proposed, which tries to preserve inter-similarities between different modalities and intra-similarity within each modality. In \cite{ding2014collective}, it was proposed to map the data from different modalities into a common subspace using projections learnt from collective matrix factorization techniques. Extending the technique of collective matrix factorization, latent semantic sparse hashing \cite{zhou2014latent} exploits sparse coding to learn hash functions. In \cite{wang2015semantic}, semantic topic multimodal hashing (STMH) is proposed which generates each bit in hash code by finding whether a concept is available in the original data or not. For achieving this, it maps the learned  multimodal semantic features into a common subspace by modeling text into multiple semantic concepts and corresponding images as latent semantic concepts. Supervised cross-modal hashing algorithms have also been proposed over time. In \cite{kumar2011learning}, spectral hashing has been incorporated with the multi-view case. In \cite{zhang2014large} semantic correlations are maximized and used to embed semantic labels into the training procedure. In \cite{zhou2014kernel}, kernel-based supervised hashing for cross-view similarity search (KSH-CV) learns kernel hash functions using adaboost algorithm. To preserve the semantic similarity, \cite{wang2016semantic} uses multi-class logistic regression to project heterogeneous data into a semantic space and uses a boosting framework to learn hash functions.

The difference between our approach and methods adopted in cross-modal hashing is that other methods assume that the information about a given class is present in both (textual and image) modalities while training the algorithm to generate hash codes for images. While in our case information in one of the mode (image) for unknown classes is absent during the training phase.

\section{Proposed Method}

Our methodology to produce hash codes for the images belonging to seen and unseen classes has been explained in this section. In section 3.1, we introduce notations that have been used throughout the paper. In section 3.2 and 3.3, we propose the approach to hash images belonging to seen classes. In section 3.4 and 3.5 we discuss the approach to hash images of unseen classes.

\subsection{Notations}

In this paper, we denote the number of seen classes by $n_{s}$ and the number of unseen classes by $n_{u}$. Let $N_{s}$ and $N_{u}$ denote the number of instances belonging to seen classes available to us during training and instances of unseen classes, respectively. Vector and its transpose  are denoted by lower case bold Roman letters such as $\mathbf{x}$ and $\mathbf{x}^\top$ respectively. Uppercase bold roman letters, such as $\mathbf{M}$, denote matrices. $\mathds{1}$ represents the indicator function. $|| \cdot ||^{2}_{F}$ represents the Frobenius norm.

\subsection{Creating anchor points}

In the feature space, ideally the set of classes must form separate clusters such that the data points belonging to a certain class should belong to the cluster representing that class. Initially, the information of only seen classes are available to us. Thus our objective is to create clusters using these training instances which represent seen classes. This objective could be formulated in the same way as that of $k$-means clustering but with a little modification that we assign a penalty of $\beta$ if the assigned cluster number for a particular image is different from its true label. The formulated objective function is shown in Eq.\ref{equation1}:

\begin{equation} \label{equation1}
\operatornamewithlimits{arg min}_{\mathbf{\Pi, \mu_{1}},...,\mathbf{\mu_{k}}}      \sum_{n,k} \pi_{nk}|| \mathbf{x_{n}} - \mathbf{\mu_{k}} || + \beta \sum_{n=1}^{N_{s}} \mathds{1} (\mathbf{\pi_{n}}\neq \mathbf{y_{n})}
\end{equation}

where $\mathbf{\mu_{i}}'s$ are cluster centers and $\Pi = [\mathbf{\pi_{1}},\ldots,\mathbf{\pi_{N_{_{s}}}}] $ is cluster assignments in one-of-$K$ encoding format. This formulation is similar to the one given in \cite{shojaee2016semi}. Though, in their formulation, the authors had assumed that the number of unseen classes were available initially and the number of clusters they assigned were equal to $ n_{s} + n_{u}$. We have not assumed that constraint and have chosen the number of clusters $k$ to be equal to the number of unseen classes $n_{s}$. $\pi_{nk}$ is equal to one, if the $n$th instance belongs to the $k$th cluster. In our experiments, we chose $\beta$ = 0.9.

Exploiting EM algorithm, $\mathbf{\mu_{i}}'s$ and $\Pi$ are updated iteratively and alternatively by optimizing the objective function. At each iteration, $\mathbf{\mu_{i}}'s$ are updated as given in Eq. \ref{equation2}

\begin{equation} \label{equation2}
\mathbf{\mu_{i}} = \frac{\displaystyle\sum_{n=1}^{n_{s}}\mathds{1}(\pi_{ni} = 1)\mathbf{x_{n}}}{\displaystyle\sum_{n=1}^{n_{s}}\mathds{1}(\pi_{ni} = 1)}
\end{equation}

$\Pi$ is updated by assigning each instance to the cluster that minimizes the corresponding term. $\mathbf{\mu_{i}}'s$ are initialized as randomly chosen data points so that they are as far as possible from each other. 

Let us call these cluster centers $\mathbf{\mu_{1}}, \mathbf{\mu_{2}},\ldots, \mathbf{\mu_{n_{s}}}$ as anchors in the rest of the paper. We also take the mean of the  features of given instances corresponding to each of the seen classes and assign the anchor to the particular class according to the Euclidean distance with respect to the mean of features corresponding to the instances belonging to that class. We will embed these in the lower dimensional space using manifold learning. The number of dimensions in the lower dimensional manifold space is equal to the length of the hash code with which we want to hash an image.

\subsection{Producing hash codes for images belonging to the seen classes}

Assuming that during training, instances correspond to only the images of seen classes, we use the cluster centers or anchors obtained by optimizing the Eq.\ref{equation1} to generate the hash codes for the images corresponding to the seen classes. Let us consider that we have a manifold-based low dimensional embedding $\mathbf{M} := \lbrace \mathbf{m_{1}},\mathbf{m_{2}},\ldots,\mathbf{m_{n_{s}}} \rbrace$ corresponding to the $n_{s}$ cluster centers. Given features of an image $\mathbf{x_{i}}$ belonging to the seen class, we aim to generate an embedding $\mathbf{m_{i}}$ such that it preserves the local neighborhood relationship both in the feature and the embedded space with respect to anchors. To obtain the embedding $\mathbf{m_{i}}$ in the manifold space, given the features $\mathbf{x_{i}}$ of an image corresponding to seen classes, the objective function shown in Eq.\ref{equation3} is minimized.
\begin{equation}\label{equation3}
O(\mathbf{m_{i}}) = \sum^{n_{s}}_{q = 1} w(\mathbf{x_{q}},\mathbf{x_{i}})||\mathbf{m_{i}} - \mathbf{m_{q}}||^{2}
\end{equation}  
where, $w(\mathbf{x_{q}},\mathbf{x_{i}})$ captures the likelihood that the given image belongs to the $q$th cluster. This can be obtained by calculating the Euclidean distance of an image from anchors $\mathbf{x_{q}}$ in the feature space. These distances from the cluster centers are then converted into probabilities or weights to which cluster, a data point belongs to using the exponential function shown in Eq.\ref{equation4}.
\begin{equation}\label{equation4}
w(\mathbf{x_{q}},\mathbf{x_{i}}) = exp \Big(\frac{-||\mathbf{x_{i}} - \mathbf{x_{q}}||^{2}}{\sigma^{2}}\Big)
\end{equation}  
where, $\mathbf{x_{i}}$ are features of $i$th image and $\mathbf{x_{q}}$ is the $q$th anchor. Here, $\sigma$ is a parameter. Differentiating $O(\mathbf{m_{i}})$ with respect to $\mathbf{m_{i}}$ and equating it to zero, we obtain equation \ref{equation6}, 
\begin{equation}\label{equation5}
\frac{\partial O(\mathbf{m_{i}}) }{\partial \mathbf{m_{i}}} \Bigg|_{\mathbf{m_{i}} = \mathbf{m_{i}}^{*}}   =  2\sum^{n_{s}}_{q = 1} w(\mathbf{x_{q}},\mathbf{x_{i}}) (\mathbf{m_{i}}^{*} - \mathbf{m_{q}}) = 0,
\end{equation}
\begin{equation}\label{equation6}
\mathbf{m_{i}^{*}} = \frac{\displaystyle\sum_{q = 1}^{n_{s}} w(\mathbf{x_{q}},\mathbf{x_{i}})\mathbf{m_{q}}}{\displaystyle\sum_{q=1}^{n_{s}}w(\mathbf{x_{q}},\mathbf{x_{i}})}
\end{equation}

The proposed method here has been inspired from Shen et. al. \cite{shen2013inductive}, where they have provided an inductive formulation to obtain the embedding of any point using the linear combination of the base embeddings. We take the top $s$ weights for our purpose and set the other weights to zero. We will call them as `nearest anchors` for a given data point throughout the paper. Apart from this, we multiplied the $i$th weight by an exponential factor $\omega^{-i}$. For our experiments, we took $\omega = 5$. This is done so that in the manifold space, the distance between the given data point and the cluster assigned to it gets further decreased. That is the value of the weight with the highest value is further increased. Finally, these top $s$ weights are re-normalized such that they sum to 1.\\
We use Eq.\ref{equation4} and Eq.\ref{equation6} to obtain the embedding of an image with features $\mathbf{x_{i}}$ belonging to any of the seen classes. Finally, we obtain the hash codes for the given image by binarizing the embedding as shown in Eq.\ref{equation7}.

\begin{equation}\label{equation7}
h(\mathbf{x_{i}}) = sign(\mathbf{m_{i}}^{*})
\end{equation}
where $sign(\cdot)$ is the element-wise sign function defined in Eq.\ref{equation8}.
\begin{equation}\label{equation8}
sign(k) =
    \begin{cases}
      1, & \text{if}\ k \geq 0 \\
      - 1, & \text{otherwise}
    \end{cases}
\end{equation}

\subsection{Producing anchors for the unseen classes}

\begin{figure}[htbp]
\centering
\includegraphics[scale = 0.5]{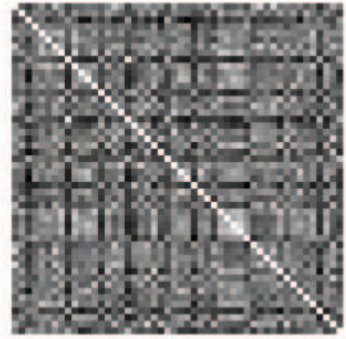}
\caption{ Cosine similarity between different classes of AwA dataset.} 
\label{fig:cosinesimilarity} 
\end{figure}

Our main concept behind producing anchors for new unseen classes is that similar classes share similar attributes. For example, the classes `MONKEY' and `CHIMPANZEE' share many attributes in common with each other, thus having high amount of similarity with each other. This could be inferred from \ref{fig:cosinesimilarity}. Thus, labels for unseen categories could be embedded using the description about how similar they are to the seen classes \cite{zhang2015zero}. We utilize the information of similarity between attributes of the classes to obtain the embeddings of the unseen classes. Thus, whenever we learn about the information (in terms of attributes) about new class by any means like textual description, we create an anchor corresponding to it. To obtain the anchor we use the cosine similarity measure between the attributes of a new class with respect to the classes, anchor of which has been obtained.

\begin{equation}\label{equation9}
w(C_{i},C_{j}) = \frac{\mathbf{a_{i}} \cdot \mathbf{a_{j}}}{||\mathbf{a_{i}}|| ||\mathbf{a_{j}}||}  
\end{equation} 

Here $C_{i}$ and $C_{j}$ are two classes between which cosine similarity is calculated and $\mathbf{a_{i}}$ and $\mathbf{a_{j}}$ are their corresponding attributes in the binary form.

We then use the Eq.\ref{equation6} to inductively obtain the embedding of the given class in the manifold space or anchor as shown in Eq.\ref{equation10}.

\begin{equation}\label{equation10}
\mathbf{m_{C_{i}}} = \frac{\displaystyle\sum_{i = 1}^{n_{s}} w({C_{q}},{C_{i}})\mathbf{m_{q}}}{\displaystyle\sum_{i=1}^{n_{s}}w({C_{q}},{C_{i}})}
\end{equation}

where, $\mathbf{m_{C_{i}}}$ is the embedding in the manifold space for the unseen class $C_{i}$. This anchor is then added to our set of base anchors. 

\subsection{Generating hash codes for images of unseen classes}

To produce the hash codes for images of unseen classes, we must embed the semantic information of different classes i.e., attributes in a common space. The framework proposed by \cite{romera2015embarrassingly} is computationally cheaper, simple and provides a closed form solution of the problem. These are the main reasons due to which we adopt their approach. 

\subsubsection{Zero-shot learning for images of unseen classes}

Let us assume that for each of the $n_{s}$ classes at training stage, we have a signature vector of size $\mathbf{a}$ such that each element of $\mathbf{a}$ lies in $[0,1]$. Signatures are represented in a matrix form as $ \mathbf{S}\in [0,1]^{a\times n_{s}} $. Let us denote all the instances available at training stage by a matrix $\mathbf{X} \in \mathbb{R}^{N_{s}\times d} $, where $d$ is the length of feature vector of each instance. All instances are labeled in one hot encoding format, with ground truth labels of each of these instances are represented using as $\mathbf{Y} \in \lbrace -1, 1 \rbrace ^{N_{s} \times n_{s}} $, with positive entry indicating the class to which the instance belongs to. To learn a predictor corresponding to the $n_{s}$ training classes, the  following objective function is optimized:
\begin{equation} \label{zeroshotequation}
\operatornamewithlimits{arg min}_{{\mathbf{V}}\in \mathbb{R}^{d\times a}} L(\mathbf{XVS}, \mathbf{Y}) + \Omega (\mathbf{V})
\end{equation}

Here $\mathbf{V}\in \mathbb{R}^{d\times a}$ embeds the semantic information in the form of attributes with image features. The regularizer chosen is of the following form shown in the equation below.
\begin{equation}
\Omega (\mathbf{V; S,X}) = \gamma || \mathbf{VS} ||^{2}_{F} + \lambda || \mathbf{XV} ||^{2}_{F} + \alpha || \mathbf{V} ||^{2}_{F}
\end{equation}

The first term of the regularizer controls attribute signature so that their representations have a similar Euclidean norm on the feature space. The second term of the regularizer checks that the approach is invariant enough to be generalized to other test feature distribution by bounding the variance of representation of instances on attribute space. The third term penalises the Frobenius norm of the weight matrix to be learned.

The scalars $\gamma, \lambda$ and $\alpha $ are the hyper-parameters. If following choices are made
\begin{enumerate}
 \item $L(\mathbf{XVS},\mathbf{Y}) = ||\mathbf{XVS} - \mathbf{Y}||^{2}_{F}$ 
 \item $\alpha = \gamma \lambda $
\end{enumerate}

then a closed solution of Eq.\ref{zeroshotequation} could be obtained as follows:
\begin{equation}
\textbf{V} = (\mathbf{X}^\top \mathbf{X} + \gamma \mathbf{I})^{-1} \mathbf{X}^\top \mathbf{YS}^\top(\mathbf{SS}^\top + \lambda \mathbf{I})^{-1}
\end{equation}

At the testing stage, we are provided with signatures $\mathbf{S}' \in [0,1]^{\mathbf{a}\times n_{u}}$ of unseen classes $n_{u}$. The probability that new instance from unseen class with feature vector $\mathbf{x_{u}}$ belongs to the $i$th class $C_{i}$ is calculated as:
\begin{equation}\label{equation13}
w(\mathbf{x_{u}},C_{i}) = \mathbf{x_{u}VS}'_{i}
\end{equation}

\subsubsection{Generating hash codes for images of unseen classes}

Once the probability to which class an image belongs to is calculated, we then use  Eq.\ref{equation13} and Eq.\ref{equation6} to inductively produce hash codes for the images belonging to the unseen classes. Here also, we take the $s$ nearest anchors for our purpose and multiplied the $ith$ weight by an exponential factor $\omega^{-i}$ before renormalizing these top $s$ weights. The manifold embedding $\mathbf{m_{x_{u}}}$ of any instance with features $\mathbf{x_{u}}$ from any of the unseen class is thus calculated as:
\begin{equation}
\mathbf{m_{x_{u}}}^{*} = \frac{\displaystyle\sum_{i = 1}^{n_{u}} w(\mathbf{x_{u}},C_{i})\mathbf{m_{C_{i}}}}{\displaystyle\sum_{i=1}^{n_{u}}w(\mathbf{x_{u}},C_{i})}
\end{equation}

where $\mathbf{m_{C_{i}}}$ is the manifold embedding of the cluster center of class $C_{i}$.
To create the hash code for the given instance we binarize the hash code by taking using the $sign(\cdot) $ function.

\section{Experimental Results}

\begin{figure}[htbp]
\centering
\subfloat[]{\includegraphics[width=1.73in]{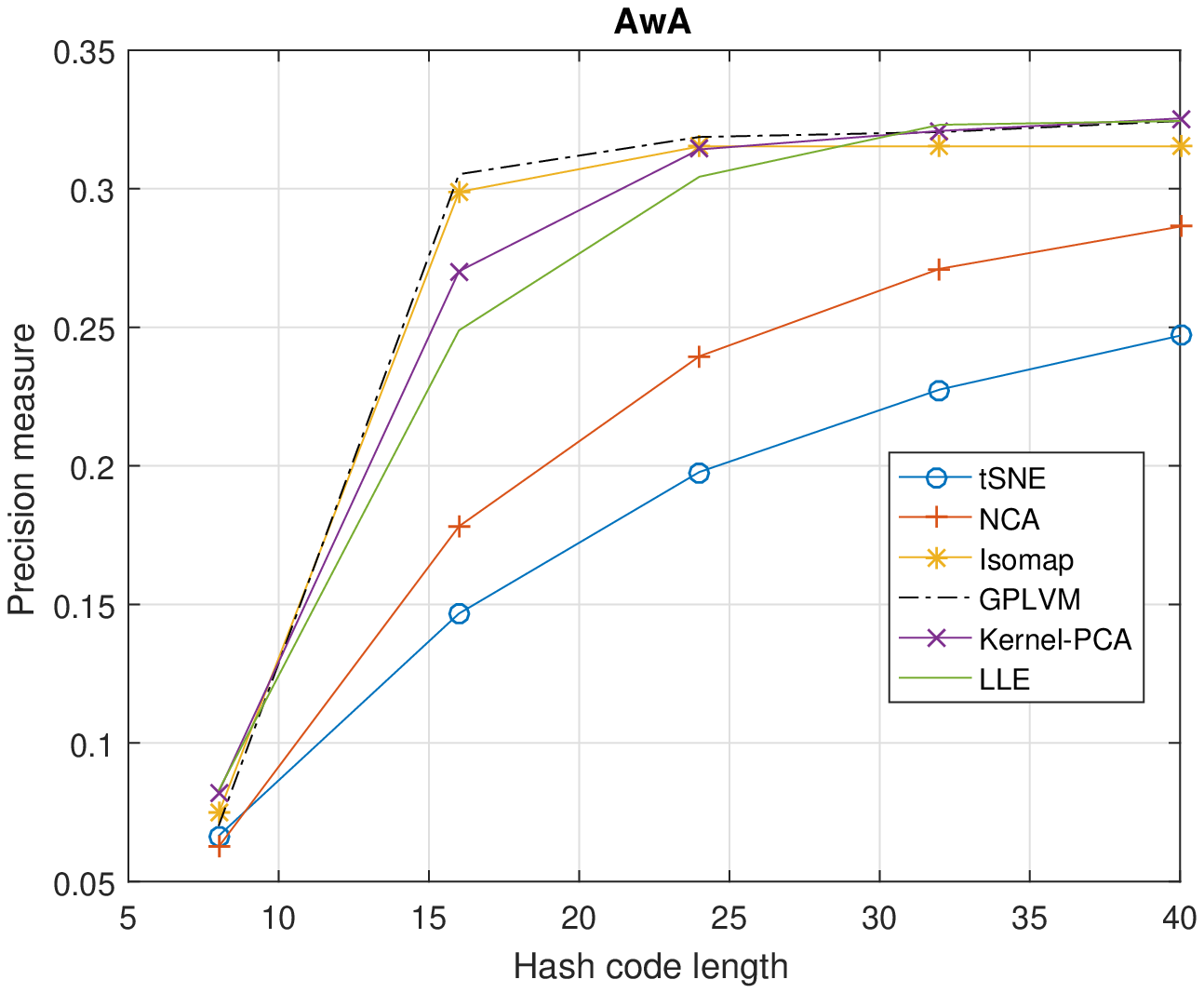}} 
\subfloat[]{\includegraphics[width=1.73in]{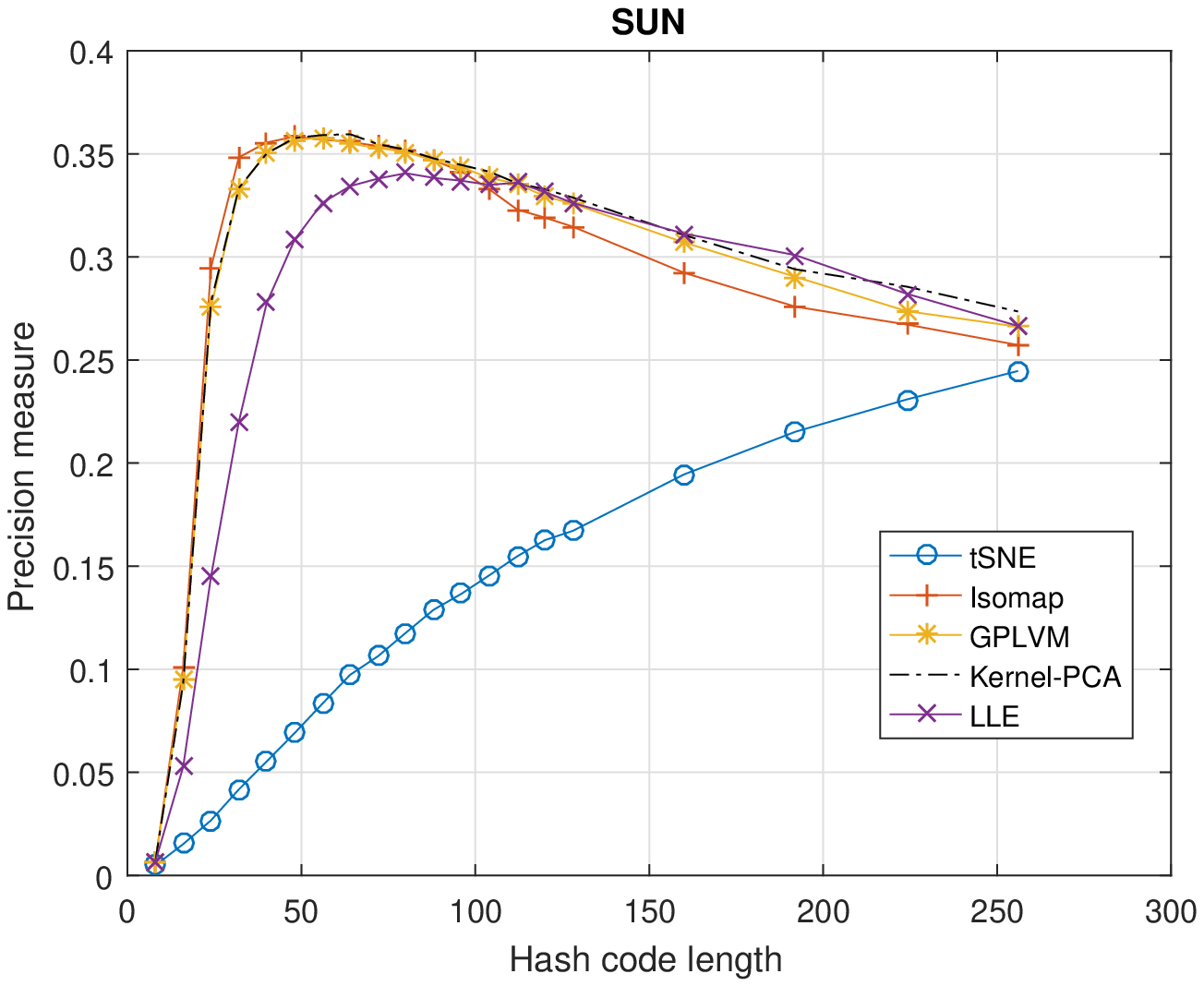}}
\caption{ Comparison of different methods on AwA (left) and SUN (right) datasets based on precision measure for varying code lengths with hamming radius 2.} 
\label{fig:precision curves} 
\end{figure}

\begin{figure}[htbp]
\centering
\subfloat[]{\includegraphics[width=1.73in]{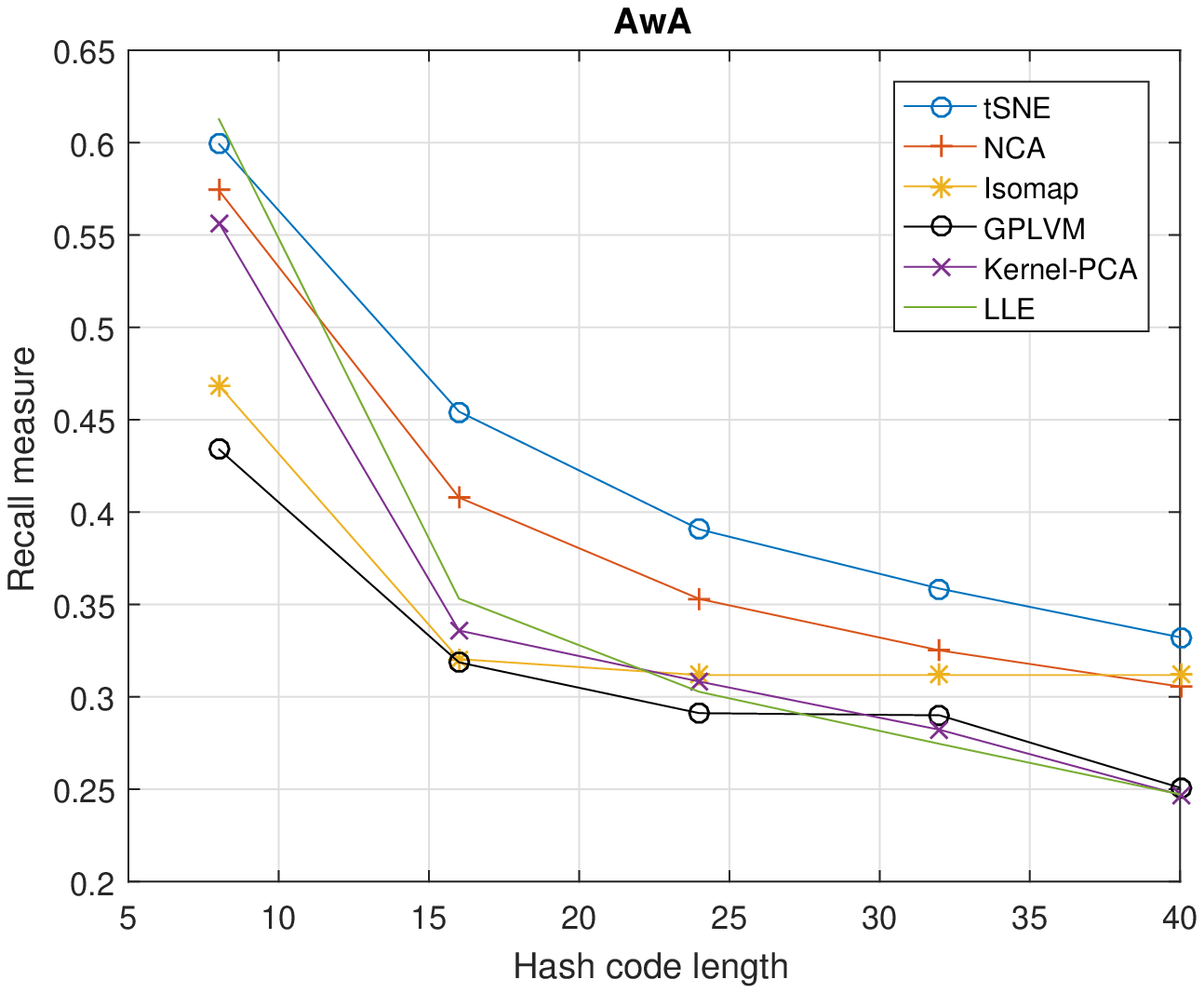}} 
\subfloat[]{\includegraphics[width=1.73in]{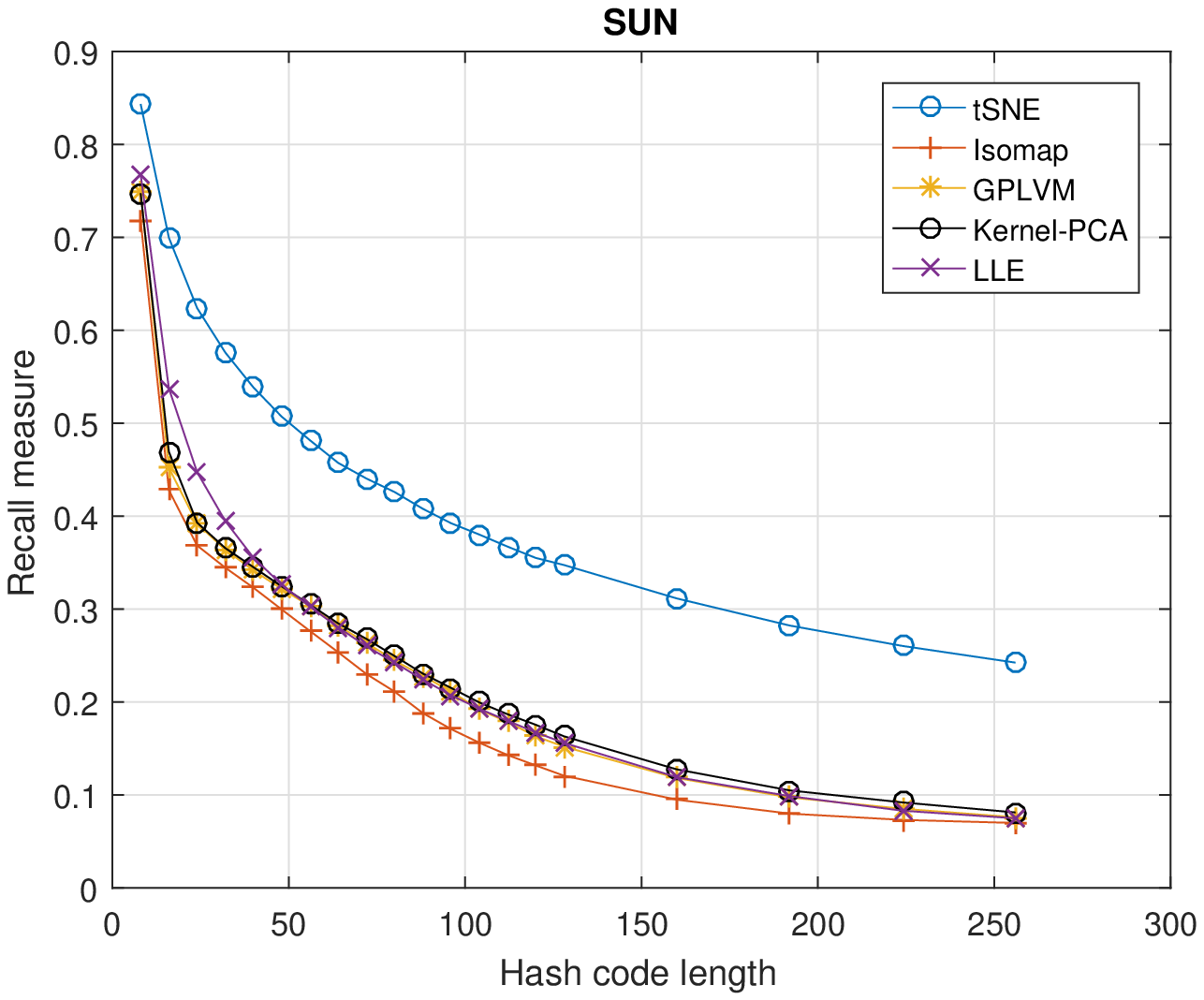}}
\caption{Comparison of different methods on AwA (left) and SUN (right) datasets based on recall measure for varying code lengths with hamming radius 2.} 
\label{fig:recall curves} 
\end{figure} 

We performed our experiments on two datasets: the SUN scene attributes dataset \cite{patterson2012sun} and the Animals with Attributes dataset (AwA) \cite{lampert2009learning}. AwA dataset consists of 30K instances. Images are categorized into one of total 50 different classes each with binary attributes of 85 dimensions. For AwA dataset, attributes are provided for each class in the dataset. We used DeCAF \cite{donahue2014decaf} features for AwA dataset. SUN dataset consists of 14,340 images each categorized into one of total 717 different classes with 20 samples for each class. Each instance has attributes of 102 dimension with value in [0,1]. For SUN dataset, attribute signature of each class is calculated by averaging the attribute signature of the instances belonging to that class. For SUN dataset, we obtained deep features using VGG with 19-layer network \cite{simonyan2014very} using MatConvNet \cite{vedaldi2015matconvnet}. We used six types of dimensionality reduction techniques in our work: Local Linear Embedding (LLE) \cite{roweis2000nonlinear}, t-Distributed Stochastic Neighbor Embedding (t-SNE) \cite{maaten2008visualizing}, Isomap \cite{tenenbaum2000global}, Kernel-PCA \cite{ham2004kernel}, Gaussian Process latent variable model (GPLVM) \cite{lawrence2004gaussian} and Neighbourhood components analysis (NCA) \cite{goldberger2004neighbourhood}. We used the matlab toolbox for dimensionality reduction \cite{van2009dimensionality} to embed anchors in manifolds. We performed two sets of experiments. For each experiment, we ran 30 trials and present the averaged results. In one set of experiments, we found the accuracy by finding the Hamming distance of the hash codes with respect to the hash code of the anchors which are generated by binarizing their embeddings. The instance was assigned to that anchor for which the Hamming distance of the instance with respect to that anchor is the lowest. We exploit the label of each image for the ground truth. In the second set of experiments, we measure the performance of our algorithm using mean of average precision (MAP), precision and recall curves. We also show the hash lookup results using F1 measure \cite{manningcambridge} which is given as follows:

\begin{displaymath}         
F1 = 2\times \frac{(precision\cdot recall)}{precision + recall}
\end{displaymath}

For evaluating F1 measure, we used Hamming distance of 2 units throughout in the above experiments. For evaluating the precision, recall and F1 measures, we divided the dataset into two sets of quarter and three-quarter size, after obtaining the hash codes of each instance in our dataset. Then for each sample in the smaller set, we find all the samples in the larger set which are at hamming distance of 2 or less than 2 units from it. We exploit the ground truths provided to us for the images in the dataset to obtain precision, recall, F1-measure and MAP measure. We used $s = 5$ in all the experiments unless specified explicitly. We also measure the time computed by each method for embedding anchors (Table.1), producing hash code for each instance in the training set (Table.2) and test set (Table.3). \\
\textbf{Results on AwA dataset}: We randomly split the dataset into training and testing part with 40 classes in our training set and rest 10 classes for testing classes. We determined the hyper parameters by randomly taking $20$ percent of the training dataset (8 classes) as our validation dataset, which were later combined with training set after fine tuning the model. For AwA dataset, the hyperparameters obtained are $\gamma = 10$ and $\lambda = 100$.  We tested our hash codes by taking hashes of bit size 8, 16, 24, 32, and 40 for AwA dataset. The maximum hash code length is equal to the number of seen classes as manifold embedding is done using the anchors which are equal to the number of seen classes. This is because, t-SNE and other manifold based techniques do not operate if the number of dimensions is less than number of data points as they are dependent on PCA framework. 

\textbf{Results on SUN dataset}: We randomly split the dataset into training and testing part with 667 classes in our training set and rest 50 classes for testing classes. We determined the hyper parameters by randomly taking $20$ percent of the training dataset (134 classes) as our validation dataset, which were later combined with training set after fine tuning the model. For the SUN dataset, the hyperparameters obtained are $\gamma = 0.01$ and $\lambda = 1$.  We tested our hash codes by taking hashes of bit size 8, 16, 24, 32, 40, 48, 56, 64, 72, 80, 88, 96, 104, 112, 120, 128, 160, 192, 224, and 256 for the SUN dataset.

\begin{table*}\label{table1}
\centering
\caption{Time (in sec) taken by each method to embed anchors in low dimensional space.}
\begin{tabular}{ |c||c|c|c||c|c|c|  }
 \hline
 
\multicolumn{1}{|c|}{\multirow{2}{*}{Method}} & \multicolumn{3}{|c|}{AwA Dataset}&\multicolumn{3}{|c|}{SUN Dataset}\\
\cline{2-7}
\multicolumn{1}{|c|}{}& 24-bit & 32-bit & 40-bit & 32-bit & 64-bit & 128-bit \\ 
 \hline

 tSNE-ZSH   & 0.117 & 0.128 & 0.128 & 6.893 & 7.198 & 8.393 \\
 NCA-ZSH    & 5.809 & 7.088 & 3.999 & 526.707  & 532.130 & 518.442 \\
 Isomap-ZSH & 0.041 & 0.038 & 0.034 & 1.307 & 1.157 & 1.138 \\
 GPLVM-ZSH  & 0.027 & \textbf{0.011} & 0.083 & \textbf{0.906} & \textbf{0.881} & \textbf{0.869} \\
 Kernel PCA-ZSH & \textbf{0.022} & 0.019 & \textbf{0.012} & 1.474 & 2.127 & 1.628 \\
  LLE-ZSH & 16.272 & 12.367 & 13.193 & 15.213 & 14.313 & 14.647 \\
 \hline
\end{tabular}
\end{table*}

\begin{table*}\label{table2}
\centering
\caption{Time (in sec) taken by each method to produce hash codes for images in the training dataset.}
\begin{tabular}{ |c||c|c|c||c|c|c|  }
 \hline

\multicolumn{1}{|c|}{\multirow{2}{*}{Method}} & \multicolumn{3}{|c|}{AwA Dataset}&\multicolumn{3}{|c|}{SUN Dataset}\\
\cline{2-7}
\multicolumn{1}{|c|}{}& 24-bit & 32-bit & 40-bit & 32-bit & 64-bit & 128-bit \\ 
 \hline
 tSNE-ZSH   & 1.271$\times 10^{-5}$ & 1.213$\times 10^{-5}$ & 1.197$\times 10^{-5}$ & \textbf{2.157}$\times$ $\bf{10^{-5}}$ & \textbf{2.184}$\times$ {$\bf{10^{-5}}$} & \textbf{2.129}$\times$ $\bf{10^{-5}}$ \\
 NCA-ZSH    & 1.056$\times 10^{-5}$ & 1.193$\times 10^{-5}$ & 1.148$\times 10^{-5}$ & 3.68$\times 10^{-5}$ & 3.536$\times 10^{-5}$ & 3.671$\times 10^{-5}$\\
 Isomap-ZSH & \textbf{9.062}$\times$ $\bf{10^{-6}}$ & \textbf{9.651}$\times$ $\bf{10^{-6}}$ & 1.102$\times 10^{-5}$ & 8.476$\times 10^{-5}$ & 8.268$\times 10^{-5}$ & 8.360$\times 10^{-5}$ \\
 GPLVM-ZSH  & 1.334$\times 10^{-5}$ & 1.143$\times 10^{-5}$ & 1.178$\times 10^{-5}$ & 9.053$\times 10^{-5}$ & 9.297$\times 10^{-5}$ & 9.104$\times 10^{-5}$ \\
 Kernel PCA-ZSH & 1.099$\times 10^{-5}$ & 1.466$\times 10^{-5}$ & \textbf{1.078}$\times$ $\bf{10^{-5}}$ & 8.452$\times 10^{-5}$ & 8.532$\times 10^{-5}$ & 8.051$\times 10^{-5}$ \\
 LLE-ZSH & 1.472$\times 10^{-5}$ & 1.152$\times 10^{-5}$ & 1.288$\times 10^{-5}$ & 3.299$\times 10^{-5}$ & 3.138$\times 10^{-5}$ & 3.287$\times 10^{-5}$ \\
 \hline
\end{tabular}
\end{table*}

\begin{table*}[ht]\label{table3}
\centering
\caption{Time (in sec) taken by each method to produce hash codes for images in the testing dataset.}
\begin{tabular}{ |c||c|c|c||c|c|c|  }
 \hline 
 \multicolumn{1}{|c|}{\multirow{2}{*}{Method}} & \multicolumn{3}{|c|}{AwA Dataset}&\multicolumn{3}{|c|}{SUN Dataset}\\
 \cline{2-7}
\multicolumn{1}{|c|}{}& 24-bit & 32-bit & 40-bit & 32-bit & 64-bit & 128-bit \\ 

 \hline
tSNE-ZSH   & \textbf{2.108}$\times$ $\bf{10^{-4}}$ & 2.118$\times 10^{-4}$ & 2.122$\times 10^{-4}$ & 7.017$\times 10^{-4}$ & 7.031$\times 10^{-4}$ & 7.019$\times 10^{-4}$\\
 NCA-ZSH    & 2.196$\times 10^{-4}$ & 2.149$\times 10^{-4}$ & 2.223$\times 10^{-4}$ & 7.918$\times 10^{-4}$ & 7.682$\times 10^{-4}$ & 7.533$\times 10^{-4}$\\
 Isomap-ZSH & 3.299$\times 10^{-4}$ & 3.209$\times 10^{-4}$ & 3.255$\times 10^{-4}$ & \textbf{6.867}$\times$ $\bf{10^{-4}}$ & \textbf{6.857}$\times$ $\bf{10^{-4}}$ & \textbf{6.832}$\times$ $\bf{10^{-4}}$\\
 GPLVM-ZSH  & 3.483$\times 10^{-4}$ & 3.418$\times 10^{-4}$ & 3.380$\times 10^{-4}$ & 8.368$\times 10^{-4}$ & 8.431$\times 10^{-4}$ & 9.284$\times 10^{-4}$\\
 Kernel-PCA-ZSH & 4.178$\times 10^{-4}$ & 4.031$\times 10^{-4}$ & 3.696$\times 10^{-4}$ & 8.443$\times 10^{-4}$ & 8.273$\times 10^{-4}$ & 8.760$\times 10^{-4}$\\
 LLE-ZSH  & 2.124$\times 10^{-4}$ & \textbf{2.092}$\times$ $\bf{10^{-4}}$ & \textbf{2.106}$\times$ $\bf{10^{-4}}$ & 8.698$\times 10^{-4}$ & 8.554$\times 10^{-4}$ & 8.756$\times 10^{-4}$\\
 \hline
\end{tabular}
\end{table*}

\begin{figure}[htbp]
\centering
\subfloat[]{\includegraphics[width=1.73in]{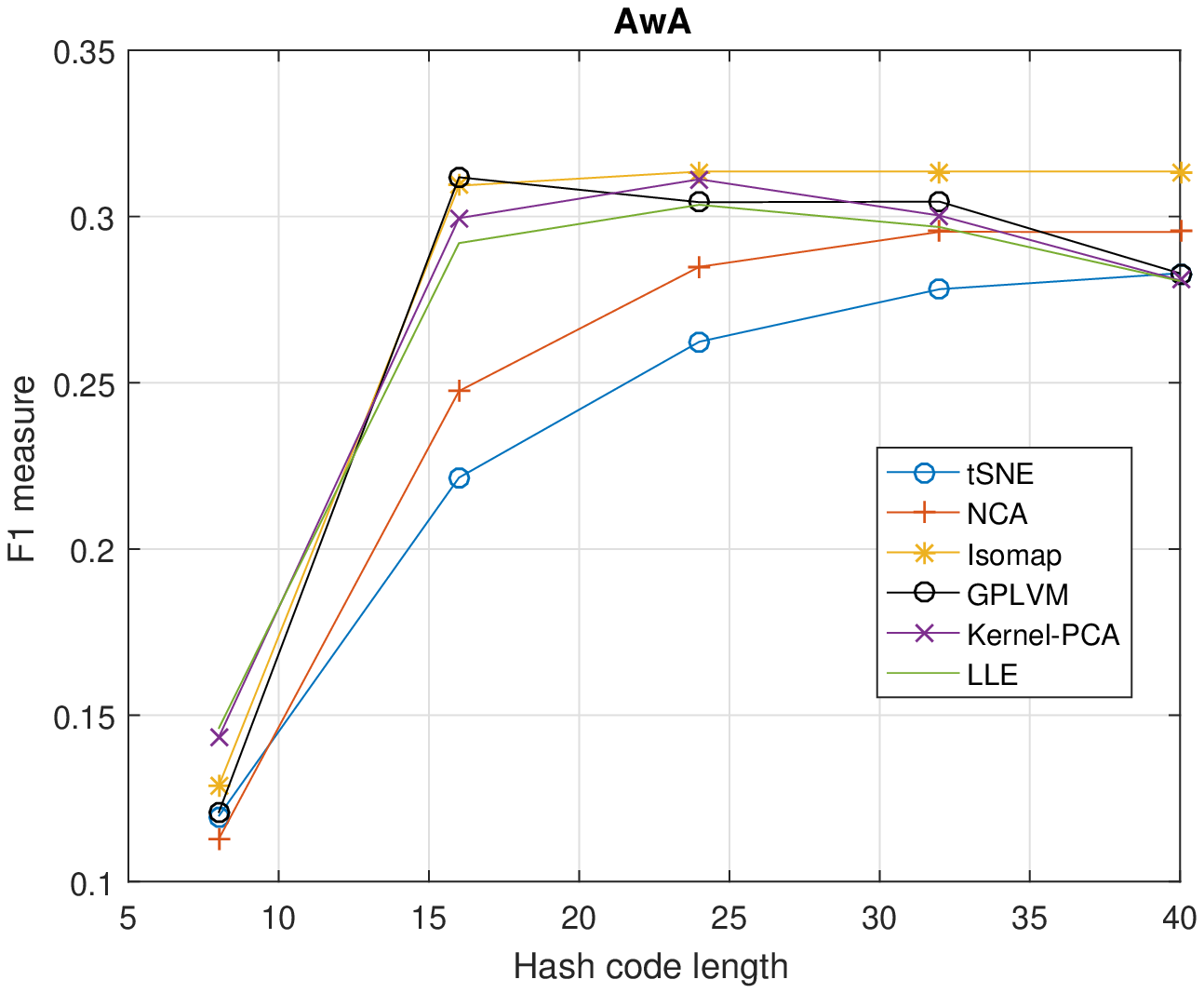}} 
\subfloat[]{\includegraphics[width=1.73in]{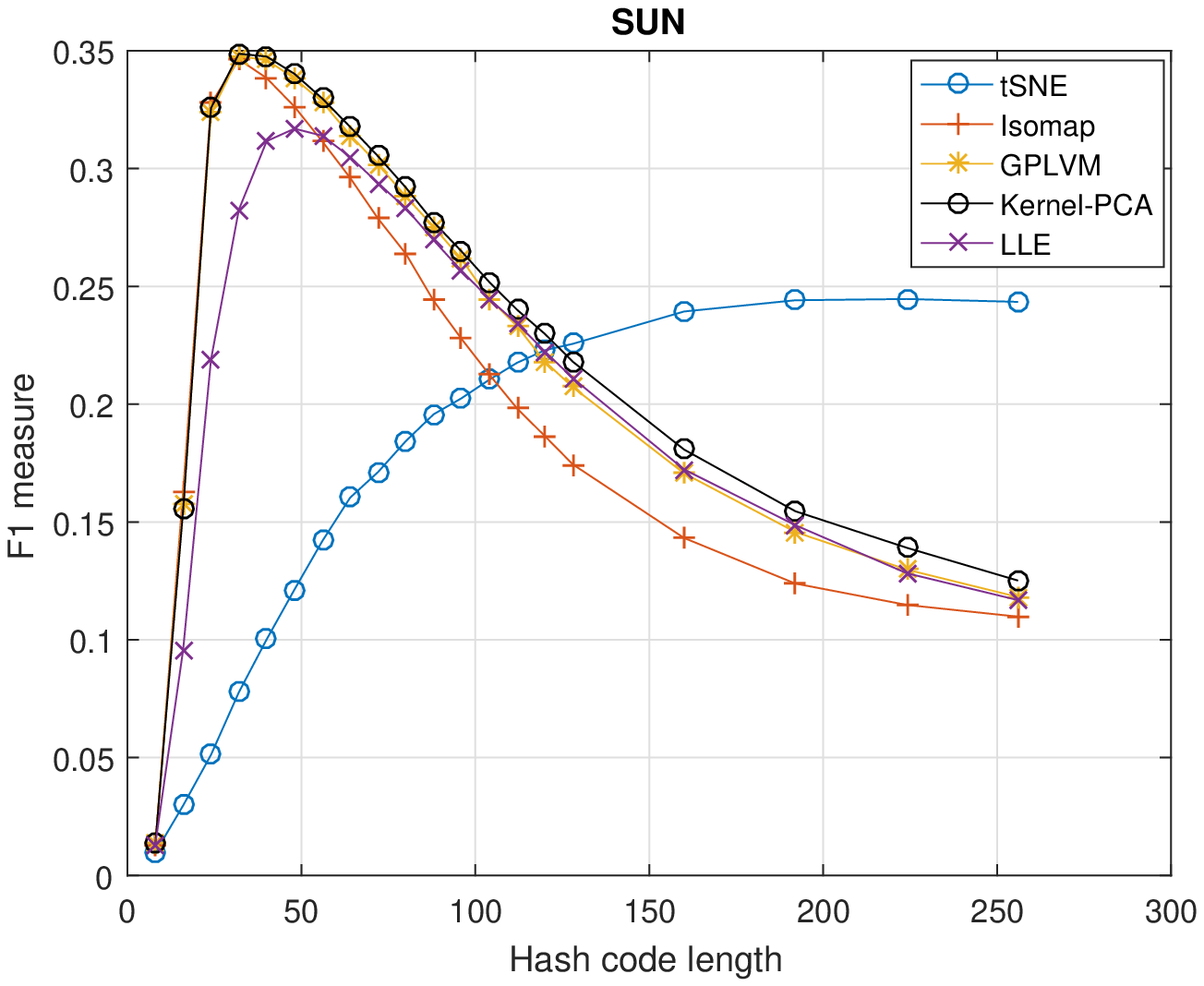}}%
\caption{Comparison of different methods on AwA (left) and SUN (right) datasets based on F1 measure for varying code lengths with hamming radius 2.} 
\label{fig:F1measure} 
\end{figure} 
\begin{figure}[htbp]
\centering
\subfloat[]{\includegraphics[width=1.73in]{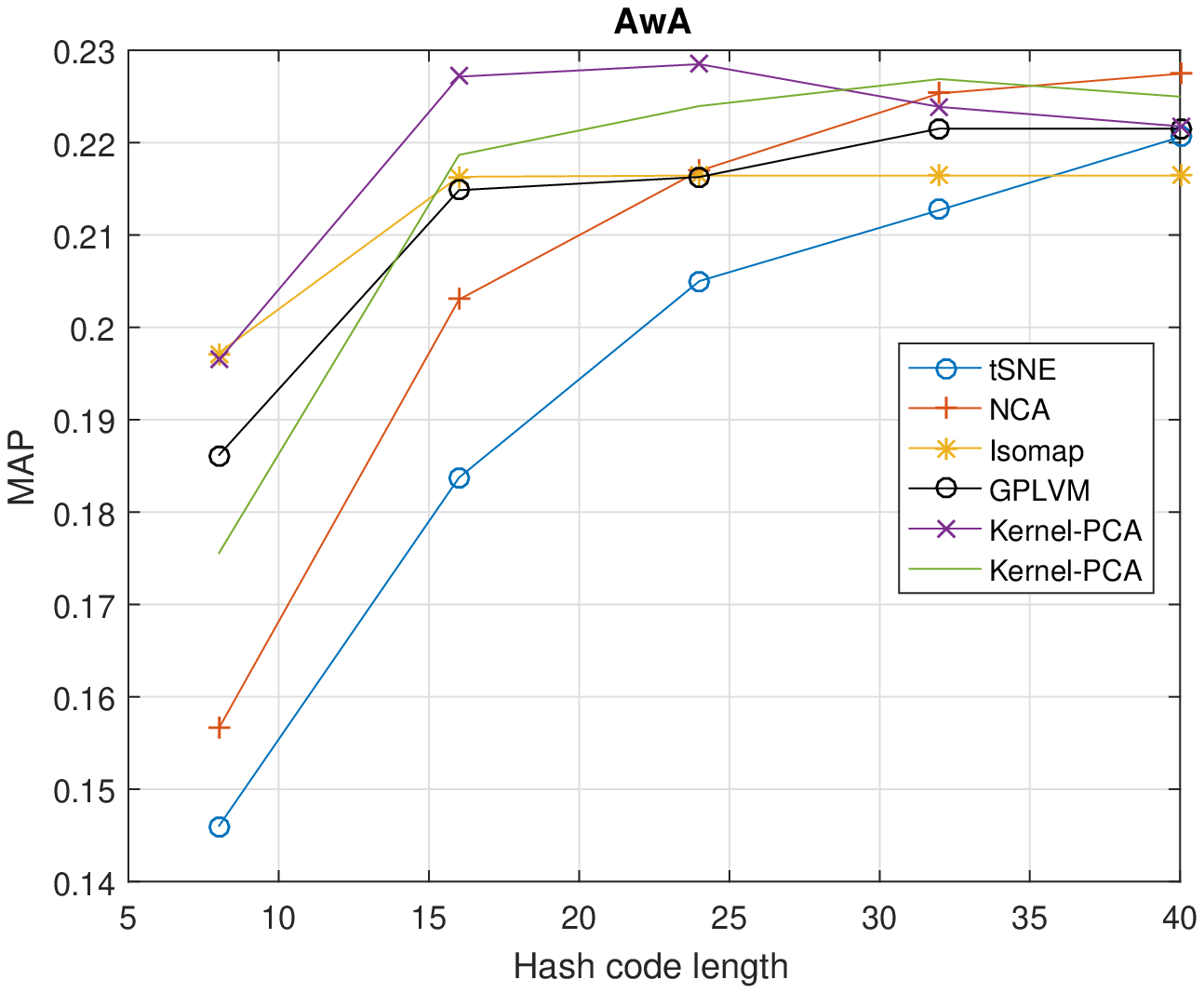}} 
\subfloat[]{\includegraphics[width=1.73in]{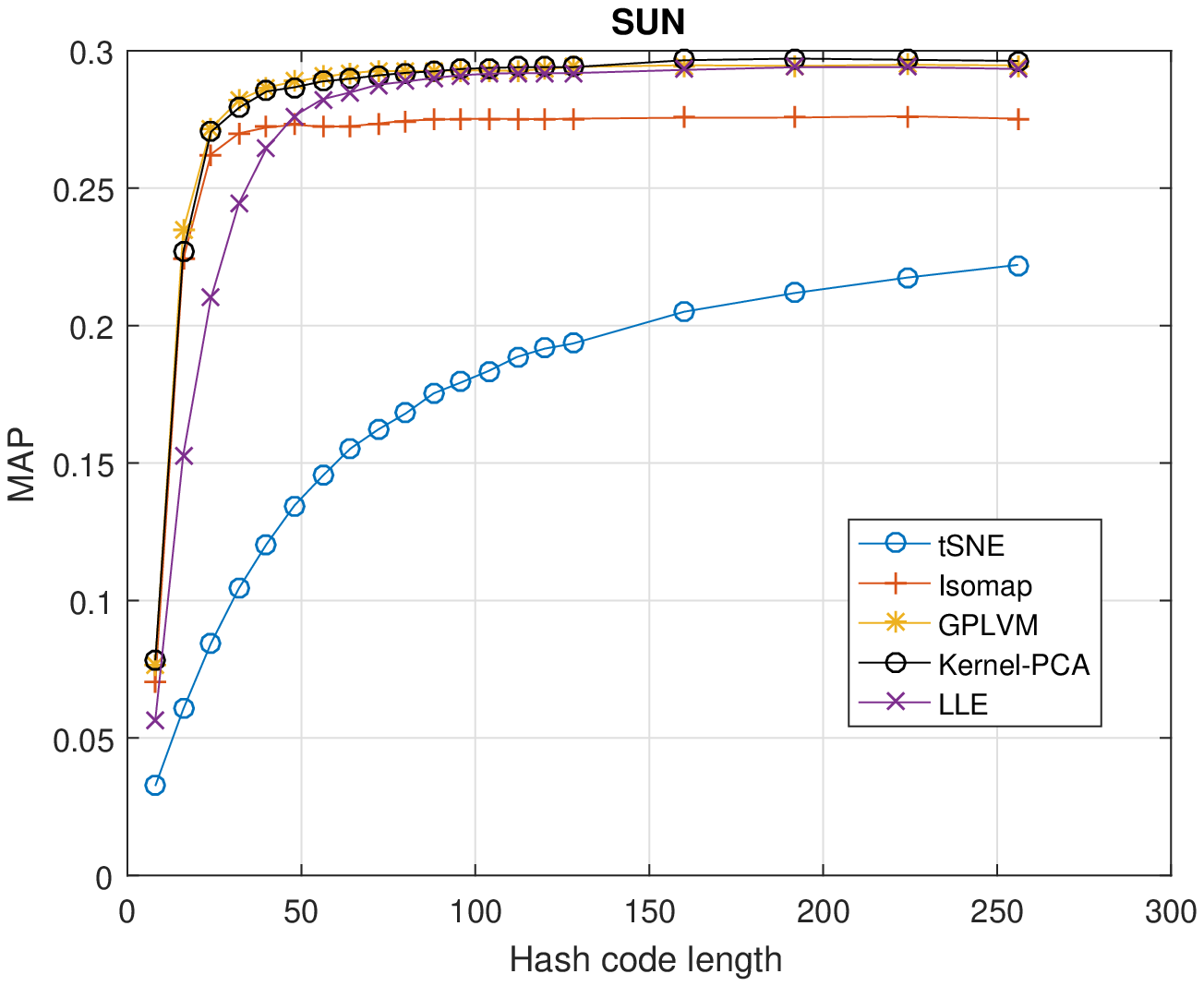}}%
\caption{Comparison of different methods on AwA (left) and SUN (right) datasets based on MAP for varying code lengths with hamming radius 2.} 
\label{fig:MAP curve} 
\end{figure} 

From Fig. \ref{fig:MAP curve}, it can be observed that for AwA dataset, with increasing hash code length, MAP measure increases for all the embeddings. But for the SUN dataset, as the hash code length increases, MAP measure for LLE based hashing decreases. For both the datasets, we observe that Kernel-PCA and NCA embedding based hashing methods perform better than the other methods. t-SNE embedding based hashing method gives poorer results when the hash code length is small but its performance significantly improves as the code length increases. From Fig.\ref{fig:precision curves}, it can be observed that Kernel PCA and GPLVM based embedding methods perform superior than the other techniques for both datasets in term of precision for the hash codes of small length. From the precision measure of  the SUN dataset, it can be observed that precision of methods except t-SNE decreases after the hash code length of 56 but it increases for t-SNE continuously with increasing code length for both the datasets. 

\begin{figure}[htbp]
\centering
\subfloat[]{\includegraphics[width=1.73in]{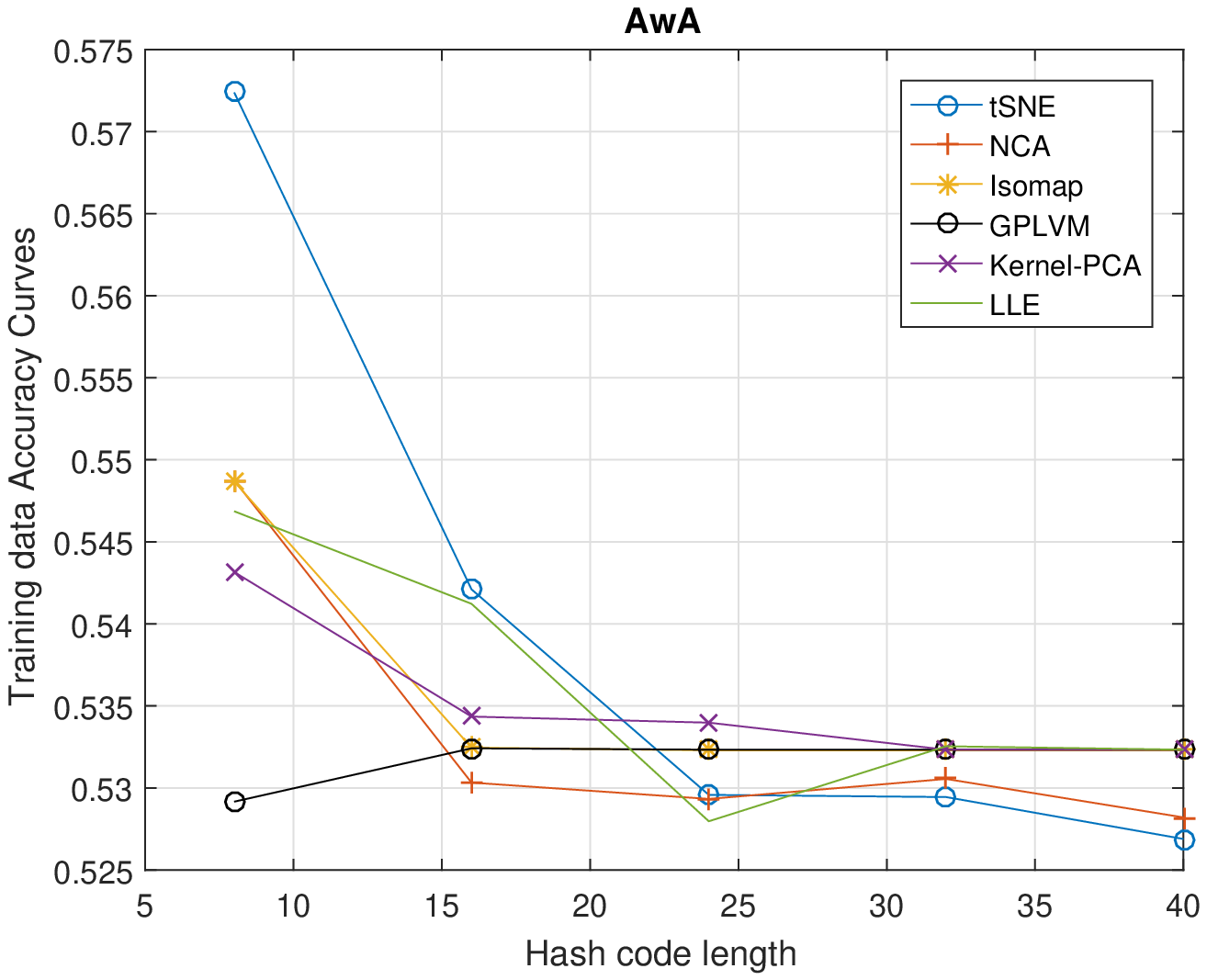}} 
\subfloat[]{\includegraphics[width=1.73in]{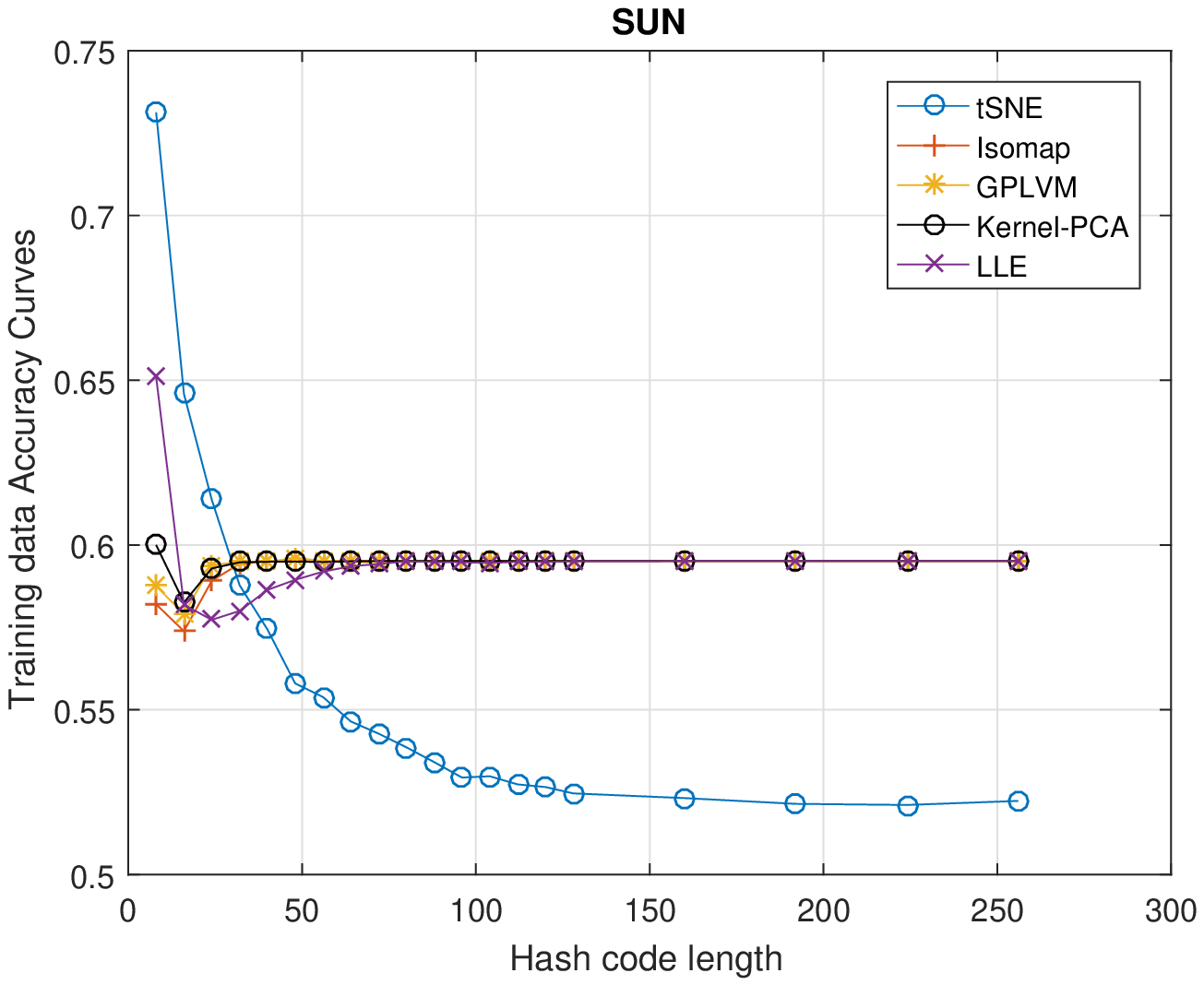}}%
\caption{Comparison of different methods on AwA (left) and SUN (right) datasets based on training data accuracy for varying code lengths with hamming radius 2.} 
\label{fig:trainingdataaccuracy} 
\end{figure}

\begin{figure}[htbp]
\centering
\subfloat[]{\includegraphics[width=1.73in]{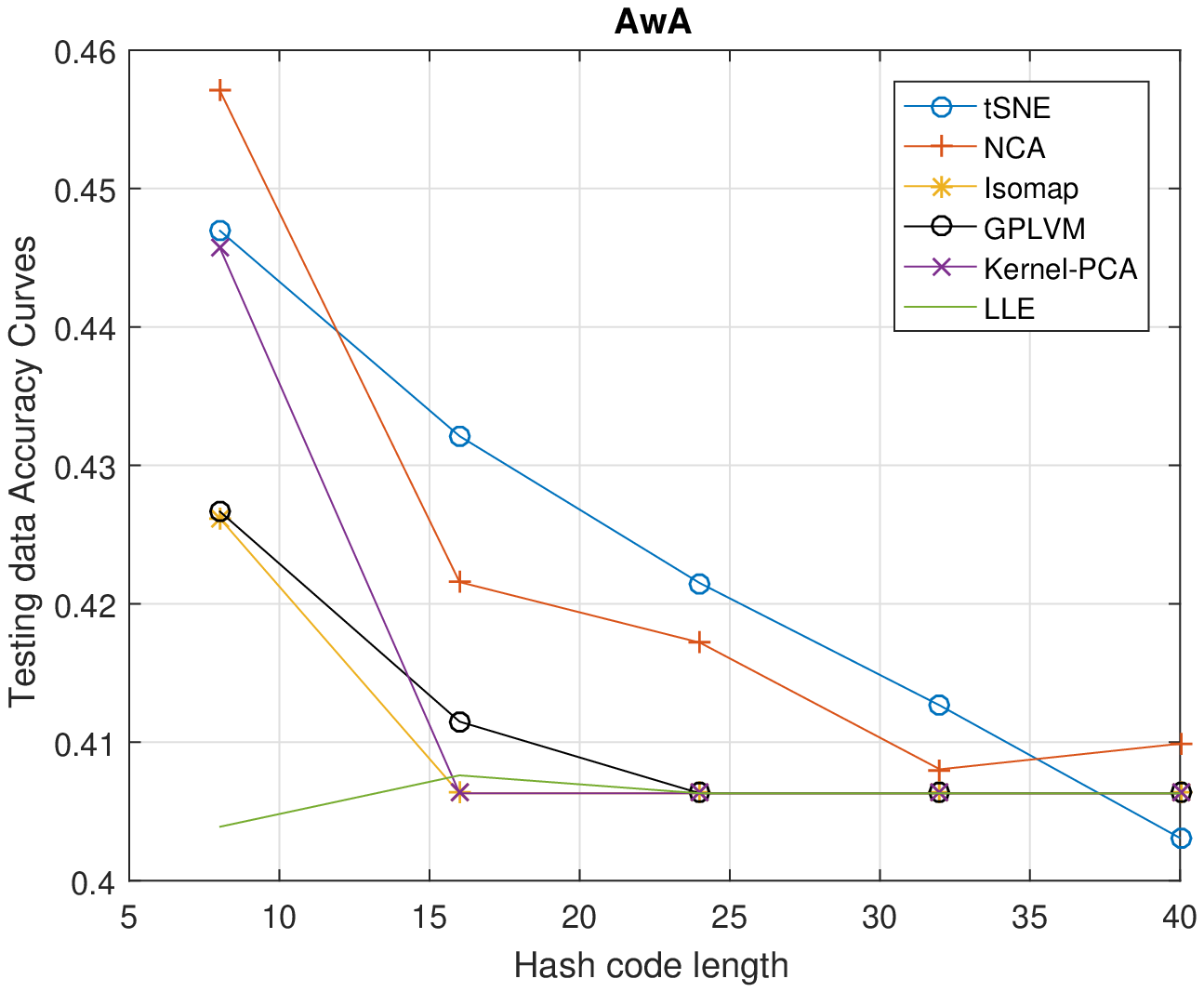}}
\subfloat[]{\includegraphics[width=1.73in]{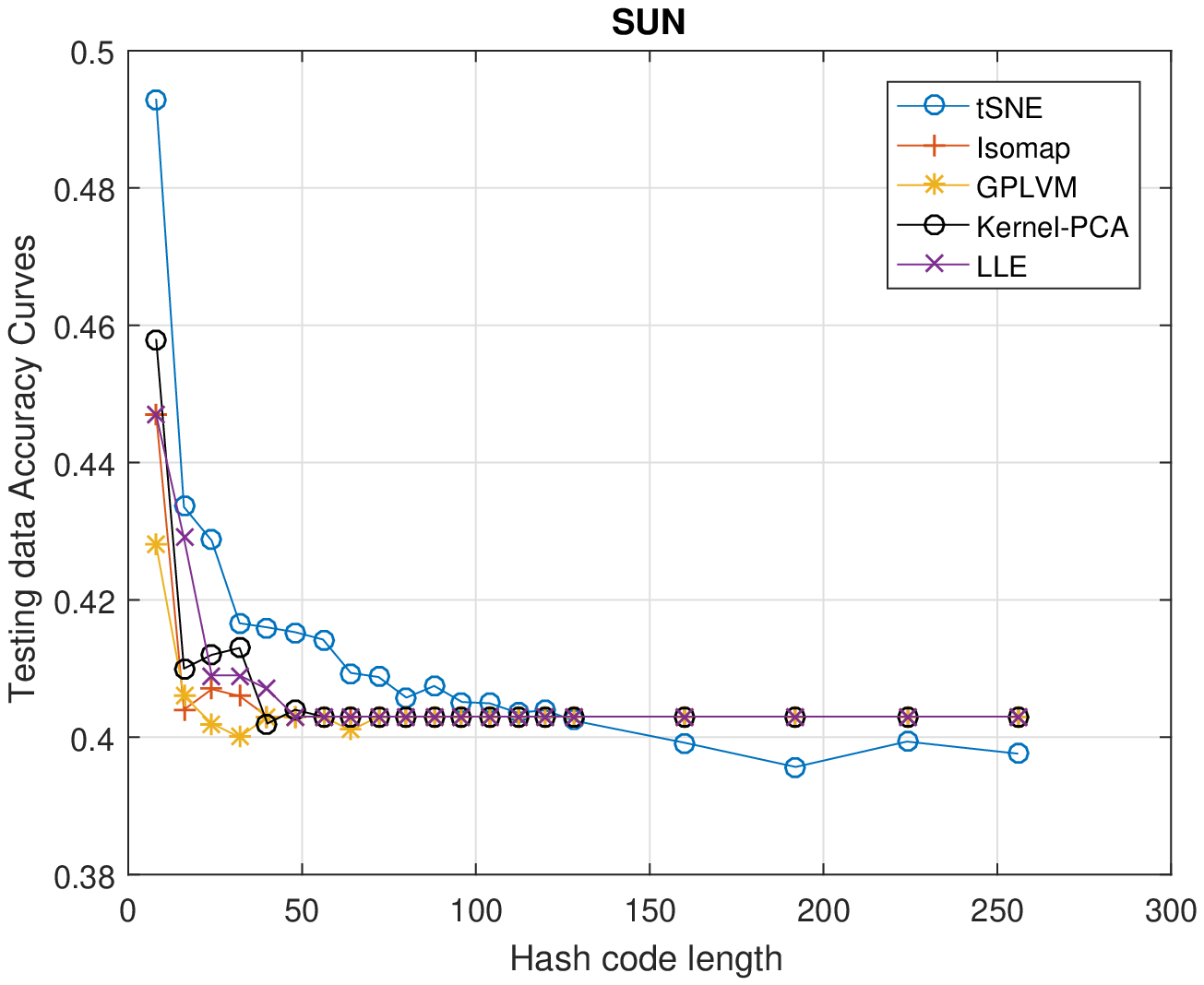}}
\caption{Comparison of different methods on AwA (left) and SUN (right) datasets based on testing data accuracy for varying code lengths with hamming radius 2.} 
\label{fig:testingdataaccuracy} 
\end{figure}

We also observe from the Fig. \ref{fig:trainingdataaccuracy}, that for AwA dataset, the accuracy for training data decreases with increasing the hash code length. While for the SUN dataset, the training data initially decreases but increases again and then saturates. For t-SNE based embedding, we observe that the training data accuracy decreases continuously. A possible reason could be that with increasing the dimension of t-SNE, it becomes less discrete and thus the hash code it generates becomes less distinct with increase in the length of hash code and hence many instances share similar hash codes. That is why, its recall is higher than other methods as it can be seen from Fig. \ref{fig:recall curves} while its precision rate being low (Fig. \ref{fig:precision curves}). Similar reasons could be given regarding the performance of our method on accuracy of instances belonging to testing classes.  The reason for this dramatic decrease in the performance of hash look up in Fig.\ref{fig:F1measure}, Fig.\ref{fig:trainingdataaccuracy} and Fig.\ref{fig:testingdataaccuracy} is that hamming spaces become sparser as we increase the hash code length.      

\begin{figure}[htbp]
\centering
\subfloat[]{\includegraphics[width=1.73in]{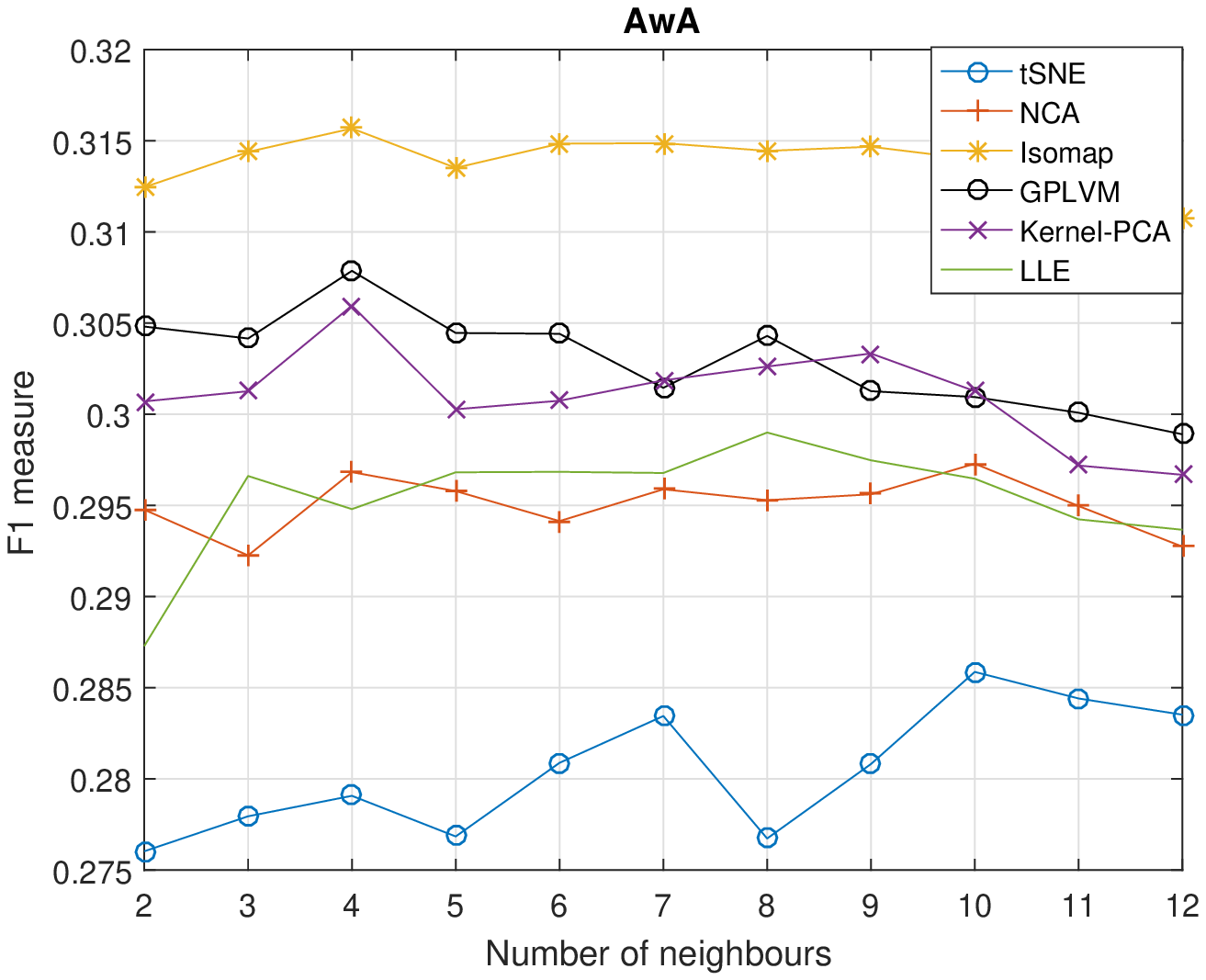}} 
\subfloat[]{\includegraphics[width=1.73in]{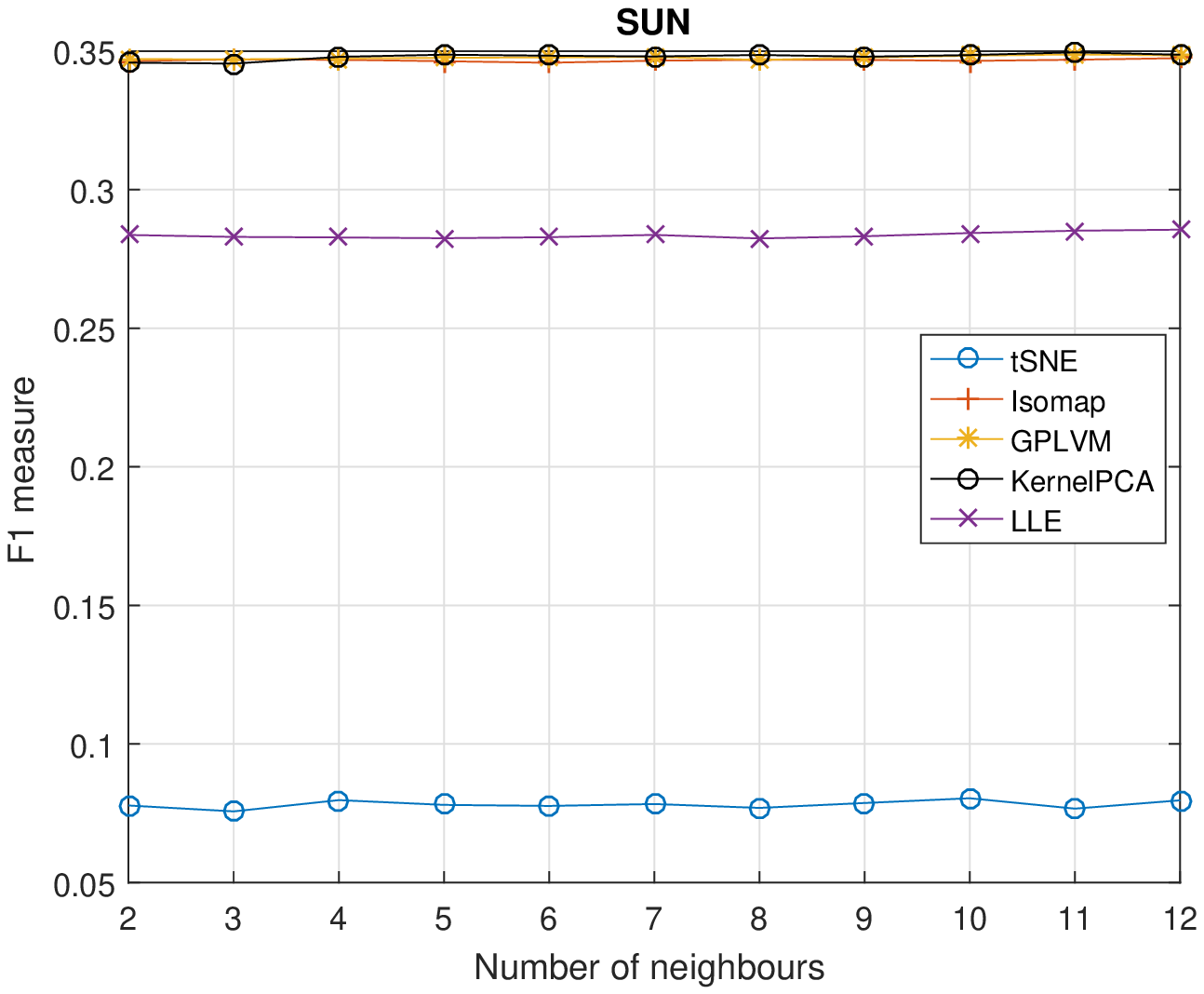}}%
\caption{Comparison of different methods on AwA (left) and SUN (right) datasets based on F1 measure for varying number of nearest anchors.} 
\label{fig:F1measure2} 
\end{figure}

\begin{figure}[htbp]
\centering
\subfloat[]{\includegraphics[width=1.73in]{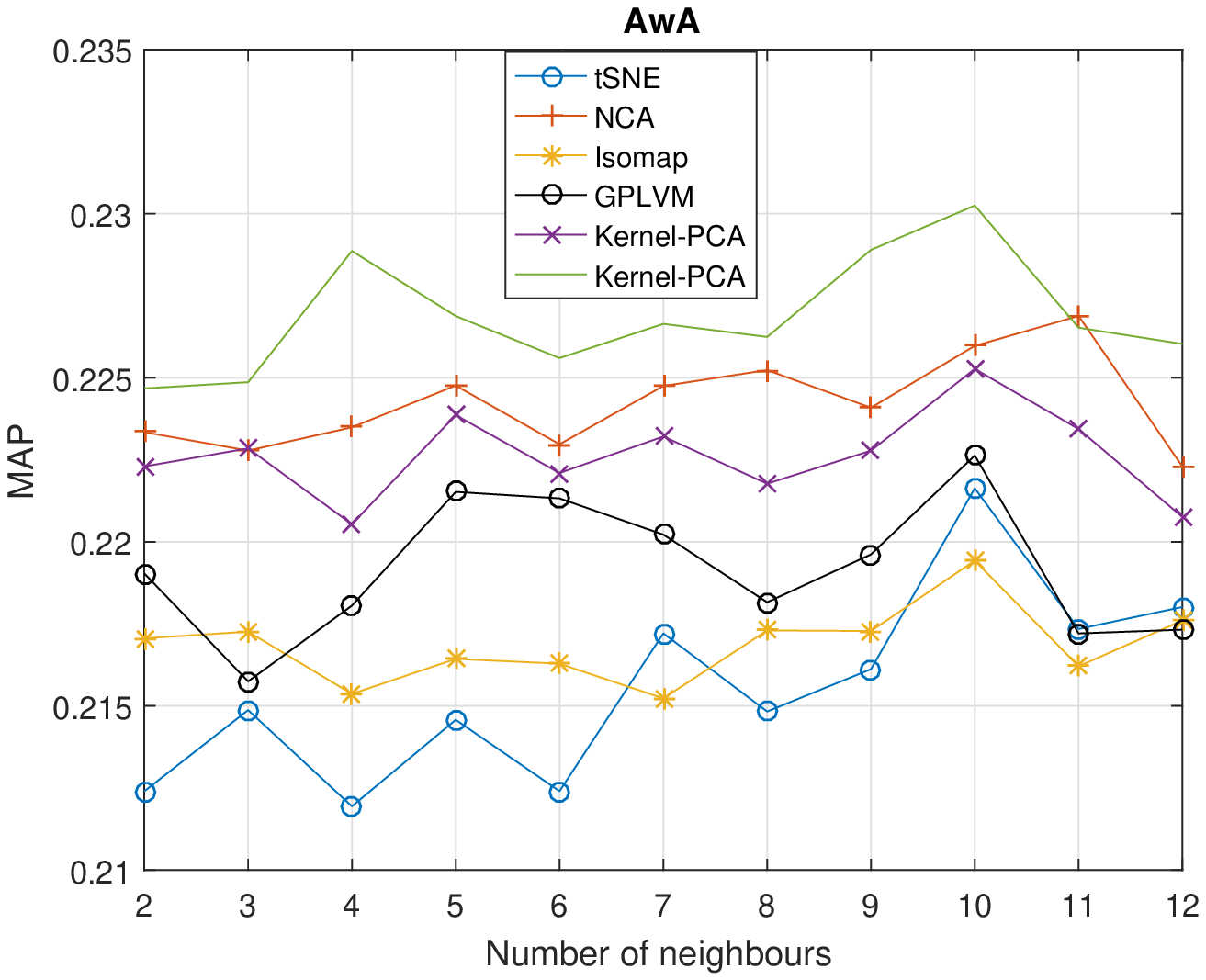}} 
\subfloat[]{\includegraphics[width=1.73in]{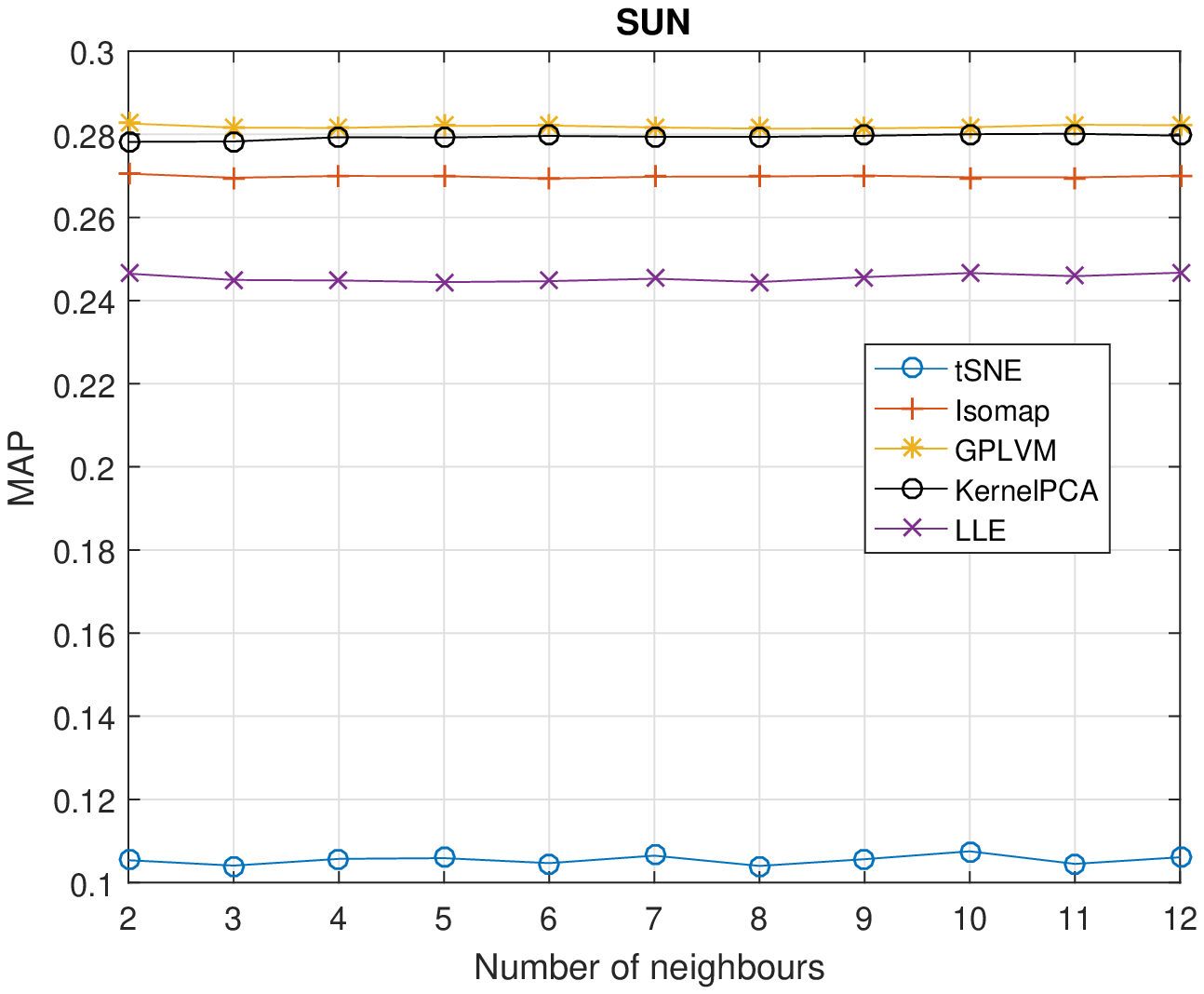}}%
\caption{Comparison of different methods on AwA (left) and SUN (right) datasets based on MAP for varying number of nearest anchors.} 
\label{fig:MAP curve2} 
\end{figure}

We also evaluate our method by varying the number of nearest anchors used to obtain the embedding of data points and plot the MAP (Fig.\ref{fig:MAP curve2}) and F1  measure (Fig.\ref{fig:F1measure2}) curves. For this, we kept the hash code length fixed to 32 bits. We can notice from the plots that the performance of the proposed methods (except t-SNE) do not change significantly by varying the number of nearest anchors for any of the manifold embedding significantly for any of the two datasets. We see that LLE based embedding ZSH algorithm performs significantly better than its counterparts for both the datasets. While t-SNE based embedding hashing technique performs poorly in terms of both F1 and MAP measure. For t-SNE embedding based hashing, its performance increases continuously as we increase the number of nearest anchors. 

\section{Conclusions}

We have proposed a hashing algorithm in the zero shot learning framework. Once the manifold embeddings are obtained corresponding to the training classes, our hashing formulation requires linear time for hashing all the training instances $(O(n))$. We used different non-parametric dimensionality reduction techniques to preserve the data distribution in the original feature space. The proposed framework exploits the information of similarity between classes to inductively generate hash codes of images belonging to the unseen classes. One advantage of this type of hashing is that if an image (for e.g. images of claws of eagle) belonging to a seen class (eagle) shares a high similarity with an unseen class (hawk), its hash code will also be similar to the instances belonging to the unseen classes and we could still retrieve that image while querying for the unseen class. This is due to the fact that in our method the anchor corresponding to the unseen class is close to the anchor corresponding to the seen class. We also provided the methodology to generate hash codes of out-of-sample data.  


\section{Future Work}

In the proposed method, we have not used significant amount of non-linearity for ranking the image features to the classes to which they belong to. In future, we could levarage deep learning methods which have achieved recently huge success in both fields of hashing and non-linear dimensionality reduction. Apart from this, deep learning techniques have achieved positive results in embedding different modalities to a common space. Moreover, recently in \cite{lei2015predicting}, authors have utilized deep learning framework for learning joint latent space. This work has inspired us to use deep networks for improving the zero shot hashing framework in the future.


\ifCLASSOPTIONcompsoc
  \section*{Acknowledgments}
\else
  \section*{Acknowledgment}
\fi

The authors would like to thank Rajendra Nagar and Aalok Gangopadhayay for helpful discussions.



%\bibliographystyle{IEEEtran}

\begin{thebibliography}{10}

\bibitem{datar2004locality}
M.~Datar, N.~Immorlica, P.~Indyk, and V.~S. Mirrokni.
\newblock Locality-sensitive hashing scheme based on p-stable distributions.
\newblock In {\em Proceedings of the twentieth annual symposium on
  Computational geometry}, pages 253--262. ACM, 2004.

\bibitem{dawkins2012illustrated}
R.~Dawkins and D.~McKean.
\newblock {\em The Illustrated Magic of Reality: How We Know What's Really
  True}.
\newblock Simon and Schuster, 2012.

\bibitem{ding2014collective}
G.~Ding, Y.~Guo, and J.~Zhou.
\newblock Collective matrix factorization hashing for multimodal data.
\newblock In {\em CVPR}. IEEE, 2014.

\bibitem{donahue2014decaf}
J.~Donahue, Y.~Jia, O.~Vinyals, J.~Hoffman, N.~Zhang, E.~Tzeng, and T.~Darrell.
\newblock Decaf: A deep convolutional activation feature for generic visual
  recognition.

\bibitem{elhoseiny2013write}
M.~Elhoseiny, B.~Saleh, and A.~Elgammal.
\newblock Write a classifier: Zero-shot learning using purely textual
  descriptions.
\newblock In {\em ICCV}. IEEE, 2013.

\bibitem{gionis1999similarity}
A.~Gionis, P.~Indyk, R.~Motwani, et~al.
\newblock Similarity search in high dimensions via hashing.
\newblock In {\em VLDB}, volume~99, pages 518--529, 1999.

\bibitem{goldberger2004neighbourhood}
J.~Goldberger, G.~E. Hinton, S.~T. Roweis, and R.~Salakhutdinov.
\newblock Neighbourhood components analysis.
\newblock In {\em NIPS}, 2004.

\bibitem{gong2013iterative}
Y.~Gong, S.~Lazebnik, A.~Gordo, and F.~Perronnin.
\newblock Iterative quantization: A procrustean approach to learning binary
  codes for large-scale image retrieval.
\newblock {\em IEEE Transactions on Pattern Analysis and Machine Intelligence},
  35(12):2916--2929, 2013.

\bibitem{ham2004kernel}
J.~Ham, D.~D. Lee, S.~Mika, and B.~Sch{\"o}lkopf.
\newblock A kernel view of the dimensionality reduction of manifolds.
\newblock In {\em ICML}. ACM, 2004.

\bibitem{kodirov2015unsupervised}
E.~Kodirov, T.~Xiang, Z.~Fu, and S.~Gong.
\newblock Unsupervised domain adaptation for zero-shot learning.
\newblock In {\em ICCV}. IEEE, 2015.

\bibitem{kulis2009learning}
B.~Kulis and T.~Darrell.
\newblock Learning to hash with binary reconstructive embeddings.
\newblock In {\em NIPS}, 2009.

\bibitem{kulis2009kernelized}
B.~Kulis and K.~Grauman.
\newblock Kernelized locality-sensitive hashing for scalable image search.
\newblock In {\em ICCV}. IEEE, 2009.

\bibitem{kulis2009fast}
B.~Kulis, P.~Jain, and K.~Grauman.
\newblock Fast similarity search for learned metrics.
\newblock {\em IEEE Transactions on Pattern Analysis and Machine Intelligence},
  31(12):2143--2157, 2009.

\bibitem{kumar2011learning}
S.~Kumar and R.~Udupa.
\newblock Learning hash functions for cross-view similarity search.
\newblock In {\em IJCAI}, 2011.

\bibitem{lampert2009learning}
C.~H. Lampert, H.~Nickisch, and S.~Harmeling.
\newblock Learning to detect unseen object classes by between-class attribute
  transfer.
\newblock In {\em CVPR}. IEEE, 2009.

\bibitem{lawrence2004gaussian}
N.~D. Lawrence.
\newblock Gaussian process latent variable models for visualisation of high
  dimensional data.
\newblock 2004.

\bibitem{lei2015predicting}
J.~Lei~Ba, K.~Swersky, S.~Fidler, et~al.
\newblock Predicting deep zero-shot convolutional neural networks using textual
  descriptions.
\newblock In {\em ICCV}. IEEE, 2015.

\bibitem{liu2014discrete}
W.~Liu, C.~Mu, S.~Kumar, and S.-F. Chang.
\newblock Discrete graph hashing.
\newblock In {\em NIPS}, 2014.

\bibitem{liu2012supervised}
W.~Liu, J.~Wang, R.~Ji, Y.-G. Jiang, and S.-F. Chang.
\newblock Supervised hashing with kernels.
\newblock In {\em Computer Vision and Pattern Recognition (CVPR), 2012 IEEE
  Conference on}, pages 2074--2081. IEEE, 2012.

\bibitem{liu2011hashing}
W.~Liu, J.~Wang, S.~Kumar, and S.-F. Chang.
\newblock Hashing with graphs.
\newblock In {\em ICML}, 2011.

\bibitem{maaten2008visualizing}
L.~v.~d. Maaten and G.~Hinton.
\newblock Visualizing data using t-sne.
\newblock {\em Journal of Machine Learning Research}, 9(Nov):2579--2605, 2008.

\bibitem{manningcambridge}
C.~D. Manning, P.~Raghavan, and H.~Sch{\"u}tze.
\newblock Cambridge university press; 2008.
\newblock {\em Introduction to Information Retrieval}, pages 158--163.

\bibitem{norouzi2011minimal}
M.~Norouzi and D.~M. Blei.
\newblock Minimal loss hashing for compact binary codes.
\newblock In {\em ICML}, 2011.

\bibitem{palatucci2009zero}
M.~Palatucci, D.~Pomerleau, G.~E. Hinton, and T.~M. Mitchell.
\newblock Zero-shot learning with semantic output codes.
\newblock In {\em NIPS}, 2009.

\bibitem{patterson2012sun}
G.~Patterson and J.~Hays.
\newblock Sun attribute database: Discovering, annotating, and recognizing
  scene attributes.
\newblock In {\em CVPR}. IEEE, 2012.

\bibitem{raginsky2009locality}
M.~Raginsky and S.~Lazebnik.
\newblock Locality-sensitive binary codes from shift-invariant kernels.
\newblock In {\em NIPS}, 2009.

\bibitem{romera2015embarrassingly}
B.~Romera-Paredes and P.~Torr.
\newblock An embarrassingly simple approach to zero-shot learning.
\newblock In {\em ICML}, 2015.

\bibitem{roweis2000nonlinear}
S.~T. Roweis and L.~K. Saul.
\newblock Nonlinear dimensionality reduction by locally linear embedding.
\newblock {\em Science}, 290(5500):2323--2326, 2000.

\bibitem{shen2015learning}
F.~Shen, W.~Liu, S.~Zhang, Y.~Yang, and H.~T. Shen.
\newblock Learning binary codes for maximum inner product search.
\newblock In {\em ICCV}. IEEE, 2015.

\bibitem{shen2013inductive}
F.~Shen, C.~Shen, Q.~Shi, A.~Van Den~Hengel, and Z.~Tang.
\newblock Inductive hashing on manifolds.
\newblock In {\em CVPR}, 2013.

\bibitem{shojaee2016semi}
S.~M. Shojaee and M.~S. Baghshah.
\newblock Semi-supervised zero-shot learning by a clustering-based approach.
\newblock {\em arXiv preprint arXiv:1605.09016}, 2016.

\bibitem{simonyan2014very}
K.~Simonyan and A.~Zisserman.
\newblock Very deep convolutional networks for large-scale image recognition.
\newblock {\em arXiv preprint arXiv:1409.1556}, 2014.

\bibitem{socher2013zero}
R.~Socher, M.~Ganjoo, C.~D. Manning, and A.~Ng.
\newblock Zero-shot learning through cross-modal transfer.
\newblock In {\em NIPS}, 2013.

\bibitem{tenenbaum2000global}
J.~B. Tenenbaum, V.~De~Silva, and J.~C. Langford.
\newblock A global geometric framework for nonlinear dimensionality reduction.
\newblock {\em science}, 290(5500):2319--2323, 2000.

\bibitem{van2009dimensionality}
L.~Van Der~Maaten, E.~Postma, and J.~Van~den Herik.
\newblock Dimensionality reduction: a comparative.
\newblock {\em J Mach Learn Res}, 10:66--71, 2009.

\bibitem{vedaldi2015matconvnet}
A.~Vedaldi and K.~Lenc.
\newblock Matconvnet: Convolutional neural networks for matlab.
\newblock In {\em Proceedings of the 23rd ACM international conference on
  Multimedia}, pages 689--692. ACM, 2015.

\bibitem{wang2015semantic}
D.~Wang, X.~Gao, X.~Wang, and L.~He.
\newblock Semantic topic multimodal hashing for cross-media retrieval.
\newblock In {\em IJCAI}, 2015.

\bibitem{wang2012semi}
J.~Wang, S.~Kumar, and S.-F. Chang.
\newblock Semi-supervised hashing for large-scale search.
\newblock {\em IEEE Transactions on Pattern Analysis and Machine Intelligence},
  34(12):2393--2406, 2012.

\bibitem{wang2014hashing}
J.~Wang, H.~T. Shen, J.~Song, and J.~Ji.
\newblock Hashing for similarity search: A survey.
\newblock {\em arXiv preprint arXiv:1408.2927}, 2014.

\bibitem{wang2016semantic}
K.~Wang, J.~Tang, N.~Wang, and L.~Shao.
\newblock Semantic boosting cross-modal hashing for efficient multimedia
  retrieval.
\newblock {\em Information Sciences}, 330:199--210, 2016.

\bibitem{weiss2012multidimensional}
Y.~Weiss, R.~Fergus, and A.~Torralba.
\newblock Multidimensional spectral hashing.
\newblock In {\em ECCV}. Springer, 2012.

\bibitem{weiss2009spectral}
Y.~Weiss, A.~Torralba, and R.~Fergus.
\newblock Spectral hashing.
\newblock In {\em NIPS}, 2009.

\bibitem{xia2014supervised}
R.~Xia, Y.~Pan, H.~Lai, C.~Liu, and S.~Yan.
\newblock Supervised hashing for image retrieval via image representation
  learning.
\newblock In {\em AAAI}, volume~1, page~2, 2014.

\bibitem{yosinski2015understanding}
J.~Yosinski, J.~Clune, A.~Nguyen, T.~Fuchs, and H.~Lipson.
\newblock Understanding neural networks through deep visualization.
\newblock {\em arXiv preprint arXiv:1506.06579}, 2015.

\bibitem{zhang2014large}
D.~Zhang and W.-J. Li.
\newblock Large-scale supervised multimodal hashing with semantic correlation
  maximization.
\newblock In {\em AAAI}, volume~1, page~7, 2014.

\bibitem{zhang2015zero}
Z.~Zhang and V.~Saligrama.
\newblock Zero-shot learning via semantic similarity embedding.
\newblock In {\em ICCV}. IEEE, 2015.

\bibitem{zhou2014latent}
J.~Zhou, G.~Ding, and Y.~Guo.
\newblock Latent semantic sparse hashing for cross-modal similarity search.
\newblock In {\em Proceedings of the 37th international ACM SIGIR conference on
  Research \& development in information retrieval}, pages 415--424. ACM, 2014.

\bibitem{zhou2014kernel}
J.~Zhou, G.~Ding, Y.~Guo, Q.~Liu, and X.~Dong.
\newblock Kernel-based supervised hashing for cross-view similarity search.
\newblock In {\em ICME}. IEEE, 2014.

\bibitem{zhu2013linear}
X.~Zhu, Z.~Huang, H.~T. Shen, and X.~Zhao.
\newblock Linear cross-modal hashing for efficient multimedia search.
\newblock In {\em Proceedings of the 21st ACM international conference on
  Multimedia}, pages 143--152. ACM, 2013.

\end{thebibliography}

\end{document}